\documentclass[runningheads]{llncs}
\usepackage{makeidx}
\usepackage{bbm}
\usepackage{graphicx}
\usepackage{subcaption}
\usepackage{amsmath,amssymb} 
\usepackage{color}
\usepackage[ruled]{algorithm2e}
\usepackage{changepage}

\newcommand{\E}{\mathbb{E}}

\begin{document}
	\pagestyle{headings}
	\mainmatter

	\def\GCPR19SubNumber{77}

	\title{On the estimation of the Wasserstein distance in generative models}

	\titlerunning{On the estimation of the Wasserstein distance in generative models}
	\authorrunning{Thomas Pinetz, Daniel Soukup, Thomas Pock}
	\author{Thomas Pinetz$^1$, Daniel Soukup$^1$, Thomas Pock$^{1,2}$}
	\institute{$^1$Center for Vision, Automation and Control, Austrian Institute of Technology \\
$^2$ Institute of Computer Graphics and Vision, Graz University of Technology, Austria}

	\maketitle

\begin{abstract}
	Generative Adversarial Networks (GANs) have been used to model the underlying probability distribution of sample based datasets. GANs are notoriuos for training difficulties and their dependence on arbitrary hyperparameters. One recent improvement in GAN literature is to use the Wasserstein distance as loss function leading to Wasserstein Generative Adversarial Networks (WGANs). Using this as a basis, we show various ways in which the Wasserstein distance is estimated for the task of generative modelling. Additionally, the secrets in training such models are shown and summarized at the end of this work. Where applicable, we extend current works to different algorithms, different cost functions, and different regularization schemes to improve generative models.
\end{abstract}

\section{Introduction}

GANs~\cite{goodfellow2014generative} have been successfully applied to tasks ranging from superresolution~\cite{ledig2017photo}, denoising~\cite{chen2018image}, data generation~\cite{arjovsky2017wasserstein}, data refinement~\cite{shrivastava2017learning}, style transfer~\cite{zhu2017unpaired}, and to many more~\cite{karras2018style}. The core principle of GANs is to pit two models, most commonly Neural Networks (NNs), against each other in a game theoretic way~\cite{goodfellow2014generative}. The first NN, denoted generator, tries to fit the data distribution of a dataset $\mathcal{X}$, and the second network, denoted discriminator, learns to distinguish between generated data and real data. Both networks learn during a so called GAN game and the final output is a generator network, which fits the real data distribution. Still, the optimization dynamics of those networks are notoriously difficult and not well understood~\cite{mescheder2017numerics}, leading to survey works concluding that no work has yet consistently outperformed the original non-saturating GAN formulation~\cite{lucic2018gans}. One key theoretical advancement is, that the previosly used Jensen-Shannon divergence is ill defined in case of limited overlap~\cite{arjovsky2017towards}. One common way to cirumvent this problem is to use different loss functions like the non-saturating loss~\cite{goodfellow2014generative} or the Wasserstein distance~\cite{arjovsky2017wasserstein}. Minimizing the Wasserstein distance yields clear convergence guarantees, given that the generator network is powerful enough~\cite{arjovsky2017wasserstein}. Still current formulations of the Wasserstein GAN (WGAN) heavily dependent on the hyperparameter setting~\cite{lucic2018gans}. Our aim with this work is to explain why this is the case and what can be done to train WGANs successfully.

We review the usage of the Wasserstein distance as it is utilized in generative modelling, showcase the pitfalls of various algorithms and we propose possible alternatives. 

As summary, our contributions are as follows:

\begin{itemize}
	\item A review and overview of common WGAN algorithms and their respective limitations.
	\item A practical guide on how to apply WGANs to new datasets.
	\item An extension to the squared entropy regularization for Optimal Transport~\cite{blondel2017smooth}, by using the Bregman distance and moving the center of the regularization.
	\item An extension on the currently available approaches to ensure Lipschitz continuous discriminator networks.
\end{itemize}


The remainder of this paper is organized as follows. In Section 2, a recap of the Wasserstein distance in the context of GANs is given. Section 3 and 4 describe all the algorithms in detail. Section 5, shows our experimental results. Our findings are summarized in Section 6 and conclusions given in Section 7.

\section{Preliminaries: Wasserstein Distance}

The p-th Wasserstein distance is defined between two probability distributions $\mu, \nu$ on a metric space $(M,c)$ as follows:

\begin{equation}
W_c^p(\mu, \nu) = \inf_{\pi \in \Pi(\mu, \nu)} \Bigg\{ \int_{M \times M} c^p(x, y) d\pi(x,y)\Bigg\}^{\tfrac 1 p},
\end{equation}

where $c(x, y)$ defines the ground cost. In this work, only the $W_c^1$ distance is considered in a discrete setting. This simplifies the whole problem to the following linear program:

\begin{equation}
W_c(\mu, \nu) = \displaystyle \min_{T \in U(\mu, \nu)} \langle T, C \rangle_F,
\label{eq:originalwdist}
\end{equation}

where $U(\mu, \nu) = \{T \in \mathbb{R}^{n \times m} | T\mathbbm{1}_n = \mu, T^T \mathbbm{1}_m = \nu\}$, $C_{ij} = c(x_i, y_j)$ and $\langle T, C \rangle_F = \sum_i \sum_j T_{ij}C_{ij}$ for any function (not necessarily a distance) $c$. This optimization problem has the following dual formulation:

\begin{equation}
\begin{split}
W_c(\mu, \nu) = \displaystyle \max_{\alpha, \beta} \alpha^T \mu + \beta^T \nu \qquad
s.t. \quad \alpha_i + \beta_{j} \leq C_{ij} \qquad \forall i,j	
\end{split}	
\label{eq:dualwdist}
\end{equation}

Based on the optimality condition of linear programing an analytical solution for $\beta$ is given by $\beta_j = \max_i C_{ij} - \alpha_i \qquad \forall_j$. By replacing the dual variables $\alpha, \beta$ with functions, namely $f(x_i) = \alpha_i$ and $f^c(y_j) = \max_i C_{ij} - \alpha_i$, the following formulation is obtained:

\begin{equation}
\begin{split}
W_c(\mu, \nu) = \displaystyle \max_{f, f^c}\E_{x \sim \mu}[f(x_i)] + \E_{y \sim \nu}[f^c(y_j)], \\
s.t. \quad f(x_i) + f^c(y_j) \leq C_{ij} \qquad \forall i,j	.
\end{split}
\end{equation}

In case $c$ is a distance $d$, it has been proven in~\cite{villani2008optimal} that $f^c(x) = -f(x)$. Using that result and rearranging the constraints yields $f(x_i) - f(y_j) \leq 1\cdot d(x_i, y_j)$, which is satisfied for all functions, which have Lipschitz constant $Lip(f) \leq 1$. This establishes the Kantorovich-Rubinstein duality, which is used in WGANs:

\begin{equation}
W(\mu, \nu) = \max_{Lip(f) \leq 1} \mathbb{E}_{x \sim \mu}[f(x)] - \mathbb{E}_{w \sim \nu}[f(w)]
\label{eq:wganform}
\end{equation} 

The objective in WGANs is to leverage the Wasserstein distance to train a NN to model the underlying distribution $\nu$, given an empirical distribution $\hat \nu$. In the GAN framework the generated distribution is constructed by using a known base distribution e.g. $z \sim N(\boldsymbol{0}, \boldsymbol{I})$, and transforming $z$ using a NN with parameters $\theta$ as follows: $\mu_\theta \sim g(z; \theta)$. The parameters $\theta$ are then learned by minimizing the distance between the parametric distribution and the empiric one ($\hat \nu$) using the following loss function:

\begin{equation}
	L(\nu, \theta) = \min_\theta W_c(\hat \nu, \mu_\theta)
\end{equation}

Due to changes in the generator parameters $\theta$ during the optimization process, the Wasserstein distance problem changes and is reevaluated in each iteration. Therefore, the speed of computation is essential. In the OT literature an additional regularization term $h(x)$ is added to improve the speed of convergence, while yielding sub-optimal results~\cite{cuturi2013sinkhorn}. This results in the following formulation:

\begin{equation}
W_{c, \epsilon}(\mu, \nu) = \displaystyle \min_{T \in U(\mu, \nu)} \langle T, C \rangle_F  + \epsilon h(T)
\label{eq:wdistreq}
\end{equation}

We discriminate between two different methodologies of algorithms, namely sub-optimal fullbatch methods and stochastic methods. The following algorithms for solving the Wasserstein distance problem to learn generative models are incorporated in our work:

\begin{enumerate}
	\item Fullbatch Methods 
	\begin{enumerate}
		\item Unregularized Wasserstein Distance (Eq. (\ref{eq:originalwdist}))
		\begin{enumerate}
			\item Primal Dual Hybrid Gradient solver (PDHG; Alg. \ref{alg:pdhg}) 
		\end{enumerate}
		\item Regularized Wasserstein Distance (Eq. (\ref{eq:wdistreq})
		\begin{enumerate}
			\item Negative Entropy Regularization
			\begin{enumerate}
				\item Sinkhorn~\cite{cuturi2013sinkhorn} (Sinkhorn; Alg. \ref{alg:sinkhorn})
				\item Sinkhorn-Center~\cite{xie2018fast} (Sinhorn-Center; Alg. \ref{alg:sinkhorncenter})
			\end{enumerate}
			\item Quadratic Regularization
			\begin{enumerate}
				\item FISTA (FISTA; Alg. \ref{alg:fista})
				\item FISTA-Center (FISTA-Center; Alg. \ref{alg:fistacenter})
			\end{enumerate}
		\end{enumerate}
	\end{enumerate}
	\item Stochastic Methods (Eq. (\ref{eq:wganform}))
	\begin{enumerate}
		\item Regularized NNs:
		\begin{enumerate}
			\item WGAN with Gradient Penalty~\cite{gulrajani2017improved} (WGAN-GP)
		\end{enumerate}
		\item Constrained NNs:
		\begin{enumerate}
			\item WGAN with Spectral Normalization~\cite{miyato2018spectral} (WGAN-SN)
			\item WGAN with convolutional Spectral Normalization (WGAN-SNC) 
		\end{enumerate}
	\end{enumerate}
\end{enumerate}

The main iterations for all algorithms are detailed in the supplementary material.

\section{Fullbatch Methods}

Fullbatch estimation means taking a data-batch of size $n$ of both probability densities and solving the Wasserstein distance for this subset $(\mathcal{X}_i, \mathcal{Y}^\theta_i)$. The idea is that the estimated Wasserstein distance is representative for the entire dataset. This is done by setting the probability for each image in the batch $x_{\{1, .., n\}} \in \mathcal{X}$ to $\mu(x) = \frac 1 n$. By the optimality conditions of convex problems, the so-called transport map $T$ is recovered. $T$ is a mapping between elements in $\hat \nu$ and $\mu_\theta$ and is plugged into the following equation to learn the generative model:
\begin{equation}
	L(\theta) = \min_\theta \langle T, c(\mathcal{X}_i, \mathcal{Y}^\theta_i) \rangle_F	
\end{equation} 

Note, that it is not necessary to differentiate through the computation of $T$, due to the envelope theorem as has been noted in~\cite{salimans2018improving,xie2018fast}. 

There are two main advantages of doing the fullbatch estimation. First, the convex solvers have convergence guarantees, which are easily checked in practice~\cite{blondel2017smooth}. Second, the convergence speed is faster than with stochastic estimates~\cite{sanjabi2018convergence}. 

\subsection{Unregularized Solver}

As a baseline, a solver for the unregularized Wasserstein distance is proposed. Therefore, the Primal-Dual Hybrid Gradient (\textbf{PDHG}) method~\cite{chambolle2011first} is used. To apply the PDHG, the problem is transformed into a saddle point problem as follows:   

\begin{equation}
W_c(\mu, \nu) = \max_{\lambda_1, \lambda_2} \min_{t \geq 0} \langle \begin{bmatrix}\lambda_1 \\ \lambda_2 \end{bmatrix}, Kt \rangle_F + c^Tt - \lambda_1^T\mu - \lambda_2^T\nu,
\label{eq:pdform}
\end{equation}

where $T, C$ are reshaped to one dimensional vectors $t, c$ and the constraints $T\mathbbm{1}_n$, $T^T\mathbbm{1}_m$ are combined to $Kt$. The full computation of the saddle point formulation and the steps of the algorithms are shown in the supplementary material.

\subsection{Regularized Optimal Transport}

Generative modelling solving the unregularized Wasserstein distance problem is computationally infeasible~\cite{genevay2018learning}. However, the solution to this problem in the OT literature is to solve for the regularized Wasserstein distance instead~\cite{luise2018differential}. In this work, either negative entropy regularization or quadratic regularization are utilized, which are defined as:

\begin{equation}
	h_e(x) = -\sum_i x_i \log(x_i), \qquad h_q(x) = \frac 1 2 ||x||^2
	\label{eq:negentropy}
\end{equation}

Negative entropy regularization leads to the \textbf{Sinkhorn} algorithm~\cite{cuturi2013sinkhorn}. While the Sinkhorn algorithm converges rapidly, it also tends to be numerically unstable and only a small range of values for $\epsilon$ lead to satisfactory results. One approach to reduce instabilities is to adopt a Bregman distance\footnote[1]{The Bregman distance is defined as follows: $D_h(x, z) = h(x) - (h(z) + \langle \nabla h(z), x - z \rangle )$} based proximal regularization term:

\begin{equation}
W_c^\epsilon(\mu, \nu) = \min_{T \in U(\mu, \nu)} \langle T, C \rangle_F + \epsilon D_h(T, T^k)
\end{equation}

Xie et al.~\cite{xie2018fast} proposed to use a modified Sinkhorn-Knopp algorithm (\textbf{Sinkhorn-Center}) with the steps given in the supplementary material.

Another way to combat the numerical stability problems and blurry transport maps is to use quadratic regularization~\cite{blondel2017smooth}. By plugging the quadratic regularization into the regularized Wasserstein distance, the following dual function is obtained:

\begin{equation}
W_{c,\epsilon} = \max_{\alpha, \beta} \alpha^T \mu + \beta^T \nu - \frac 1 {2\epsilon} \sum_{ij} [\alpha_i + \beta_j - C_{ij}]_+^2.
\end{equation}

The dual problem can be directly solved by the FISTA algorithm~\cite{beck2009fast}. FISTA was chosen due to its optimal convergence guarantees for problems like this and due to its simple iterates as is shown in the supplementary material (Alg. \ref{alg:fista}). The transport map $T$ is given by: $T = \tfrac 1 \epsilon \sum_{ij} [\alpha_i + \beta_j - C_{ij}]_+$. To improve the convergence speed and allow higher values for $\epsilon$, we also consider a proximal regularized version. The cost function, whose derivation is contained in the supplementary material is:

\begin{equation}
W_c^\epsilon = \max_{\alpha, \beta} \alpha^T \mu + \beta^T \nu - \sum_j \sup_{t_j \geq 0} \langle  t_j, \alpha + \beta_j \mathbbm{1}  - C_j \rangle - \frac \epsilon 2 ||t_j - t^k_j||_2^2  	
\end{equation}

This is again solved using the FISTA algorithm in Alg. \ref{alg:fistacenter}, with $T = T^n$.

\section{Stochastic Estimation Methods}

Full batch methods rely on the option to use batches, which are indicative for the entire problem. The required batch size is enormous for large scaled tasks~\cite{peyre2019computational}. In practice, the fact that close points in the data space have similar values for their Lagrangian multipliers suggest the usage of functions, which have this property intrinsically. Therefore, the Wasserstein distance is commonly approximated with a NN. The Kantorovich-Rubinstein duality leads to a natural formulation using a NN, named $f$:

\begin{equation}
W_c(\mu, \nu_\theta) = \max_{Lip(f) \leq 1} \min_\theta \E_{x \sim \mu}[f(x)] - \E_{z \sim N(\boldsymbol{0}, \boldsymbol{I})}[f(g_\theta(z))] 
\end{equation}

The key part of this formulation is the Lipschitz constraint~\cite{qin2018gan}. In practice, one of two ways is used to ensure the Lipschitzness of a NN, which is either by adding a constraint penalization to the loss function or constraining the NN to only allow 1-Lipschitz functions.

\subsection{Lipschitz Regularization}

Here the following observation is used. If $||g|| \leq 1$ holds for $g \in \partial f$, then $Lip(f) \leq 1$. By observing this fact, a simple regularization scheme, named gradient penalty, has been proposed and is widely used in practice~\cite{gulrajani2017improved}:

\begin{equation}
\begin{split}
{}& GP(x) = \lambda(||\nabla f(x)||_2 - 1)^2 \\  
{}&\forall x \in \mathbb{R}^n \quad \exists y \in \mathcal{X} , z \in \mathbb{R}^d, \alpha \in [0, 1] \subset \mathbb{R} \quad  \text{s.t.} \quad  x = \alpha y + (1 - \alpha) g_\theta(z)
\end{split}
\label{eq:gp}
\end{equation}

One thing to note in this formulation the number of constraints is proportional to the product of the number of samples, generated images and the granularity of $\alpha$, which makes the algorithm only slowly converging.

\subsection{Lipschitz Constrained NN}

One can interpret NNs as hierarchical functions, which are composed of matrix multiplications, convolutions (also denoted as matrix multiplications) and non-linear activation functions $\sigma_i$:

\begin{equation}
	f(x) = \sigma_0(W_0(\sigma_1(W_1(... \sigma_n(W_n x)))))
\end{equation}

The Lipschitz constants of such a function can be bounded from above by the product of the Lipschitz constant of its layers:

\begin{equation}
	Lip(f) \leq \Pi_i^l Lip(f_i), 
\end{equation}

where $f_i(x) = \sigma_i(W_i x)$. Therefore, if each layer is 1-Lipschitz the entire NN is 1-Lipschitz. Common activation functions like ReLU, leaky ReLU, sigmoid, tanh, and softmax are 1-Lipschitz. Therefore, if the linear maps $W_i$ are 1-Lipschitz so is the entire network as well~\cite{gouk2018regularisation}. The Lipschitz constant of the linear maps are given by their spectral norm $||W||_2$. In the WGAN-SN algorithm, the spectral norm is computed using the power method~\cite{miyato2018spectral}. The power method (Alg. \ref{alg:pm}) converges linearly depending on the ratio of the two largest eigenvectors $\lambda_1, \lambda_2$: $O((\tfrac {|\lambda_1|} {|\lambda_2|})^k)$~\cite{gouk2018regularisation}. For matrices this is done by using simple matrix multiplications. For convolutions, in the \textbf{WGAN-SN} algorithm the filter kernels are reshaped to 2D, the power method is applied and then the result is reshaped back~\cite{miyato2018spectral}. It is trivial to construct cases, where this is arbitrarily wrong~\cite{gouk2018regularisation} and in Fig. \ref{fig:gradnorm} the deviation from $1$-Lipschitzness is demonstrated. A more detailed example is shown in the supplementary material. Therefore, a mathematically correct algorithm, namely the \textbf{WGAN-SNC} is proposed, where we apply a forward and a backward convolution onto a vector $x$ in each iteration, which actually mimics the matrix multiplication of the induced matrix by the convolution. Gouk et al.~\cite{gouk2018regularisation} proposed a similar power method for classification and projected the weights back onto the feasible set after each update step for a classification network. In the supplementary material it is shown empirically on simple examples that this is too prohibitive to estimate the Wasserstein distance reliably. Therefore, the WGAN-SNC algorithm applies power method as a projection layer, similar to the WGAN-SN algorithm. In that layer, the $u$ variable persists across update steps, an additional iteration is run during training, and the projection is used for backpropagation.

\begin{algorithm}[t]
	\SetAlgoLined
	\KwResult{The spectral normalized weight matrix $W_s$}
	$u^0 = N(\boldsymbol{0}, \boldsymbol{I})$ \\
	\For{k = $1, ...,  n$}{
		$u^{k + \tfrac 1 2} = W^TWk$ \\
		$u^{k + 1} = \frac {u^{k + \tfrac 1 2}} {||u^{k + \tfrac 1 2}||_2}	$
	}
	$W_s = \frac {W} {||Wu^n||_2}$
	\caption{Power method: Requires the matrix $W$ and number of iterations $n$ with default value $1$.}
	\label{alg:pm}
\end{algorithm}

\section{Experiments}
\label{sec:exp}

The base architecture for all the NNs in this work is a standard convolutional NN as is used by the WGAN-SN~\cite{miyato2018spectral}, which is based on the DCGAN~\cite{radford2015unsupervised}. Details are described in the supplementary material. The default optimizer is the Adam optimizer with the parameter setting from the WGAN-GP~\cite{gulrajani2017improved} setting ($lr=0.0001$, $betas=(0.0,0.9)$). We use 1 discriminator iteration for WGAN-SN(C) algorithms and 5 for WGAN-GP.

\subsection{MNIST Manifold Comparison}

\begin{figure}[t]
	\begin{adjustwidth}{-1cm}{}
	\centering
	\begin{subfigure}[b]{0.48\textwidth}
		\centering
		\includegraphics[width=0.85\textwidth]{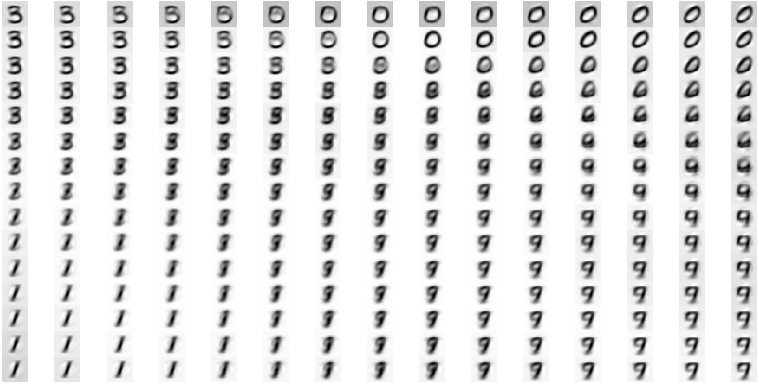}
		\caption{Cosine-distance}
	\end{subfigure}
	\begin{subfigure}[b]{0.48\textwidth}
		\centering
		\includegraphics[width=0.85\textwidth]{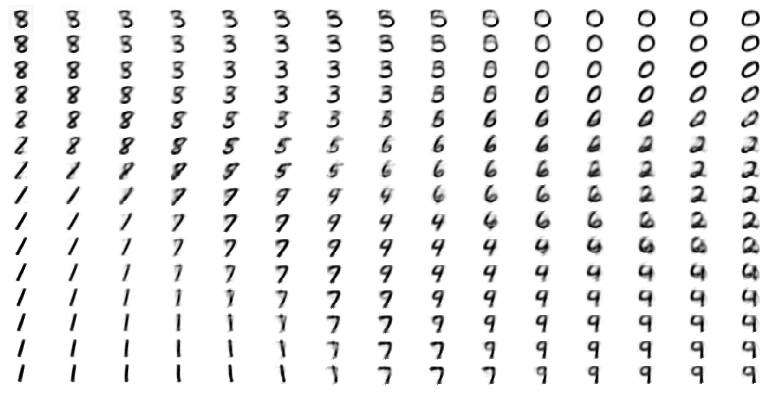}
		\caption{L2 Norm}
	\end{subfigure}
	
	\begin{subfigure}[b]{0.48\textwidth}
		\centering
		\includegraphics[width=0.85\textwidth]{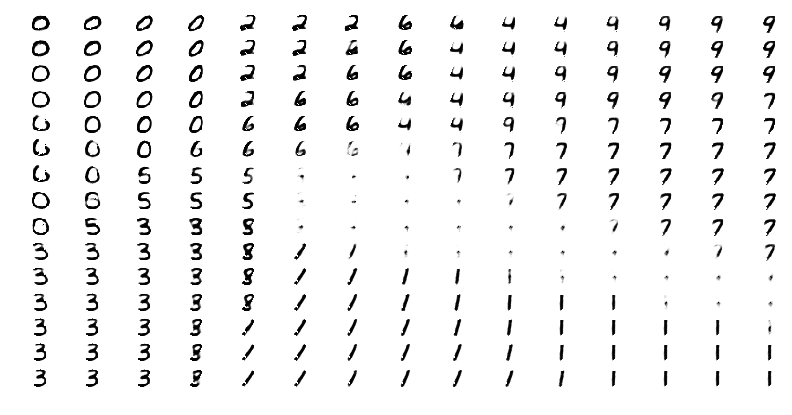}
		\caption{L1 Norm}
	\end{subfigure}
	\begin{subfigure}[b]{0.48\textwidth}
		\centering
		\includegraphics[width=0.85\textwidth]{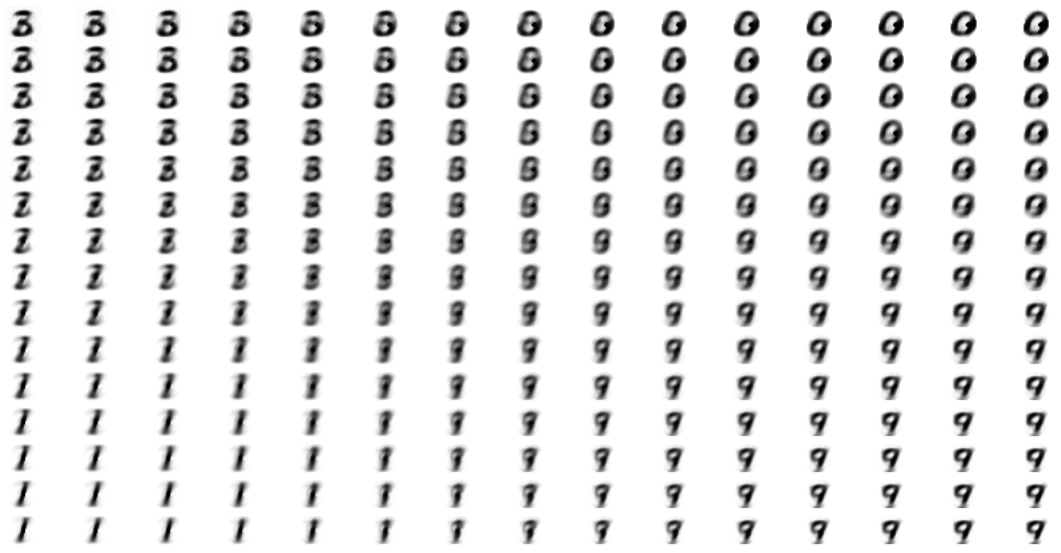}
		\caption{SSIM}
	\end{subfigure}
	\caption{Impact of different cost functions on the MNIST manifold, trained using a Sinkhorn-GAN. Notice, the different interpolations between the digits (L1 sharper, L2 blurrier) and image quality (L1 some images show no digit) and the occurrence of each digit in the manifold (SSIM is missing 2,4,5,6).}
	\label{fig:manifold}
	\end{adjustwidth}
\end{figure}

Here, the impact of the cost function on the generated manifold is investigated. The L2-norm is compared to the L1-norm, cosine-distance~\cite{salimans2018improving} and SSIM distance~\cite{wang2004image}. For this example a generator NN was trained with 1 hidden layer with 500 neurons taking $z \sim U^2(0, 1)$ as input and producing an image as output. This network is then trained using the Sinkhorn-Knopp algorithm on a batch of $1000$ samples, the manifold of which is shown in Fig. \ref{fig:manifold}. In accordance to the image processing literature, the L1 norm produces crisper images and transitions between the images then the other cost functions. However, not all the images in the manifold show digits. On the other hand the L2 norm produces digit images everywhere, similar to the output of the WGAN-GP algorithm on large datasets, but the transitions are blurry. The cosine-distance is just a normalized and squared L2-distance. Still, the resulting manifold is quite different, as it fails to capture all the digits. Also the images are blurrier than using the actual L2-norm. This leads to the conclusion that by normalizing the images, information is lost and it is harder to separate different images. While the SSIM does generate realistic digit images, it fails at capturing the entire distribution of images, e.g. digits $4$ or $6$ do not occur in the manifold. 

\subsection{Hyperparameter Dependence}

\begin{figure}[t]
	\centering
	\begin{subfigure}[b]{0.48\textwidth}
		\centering
		\includegraphics[width=\textwidth]{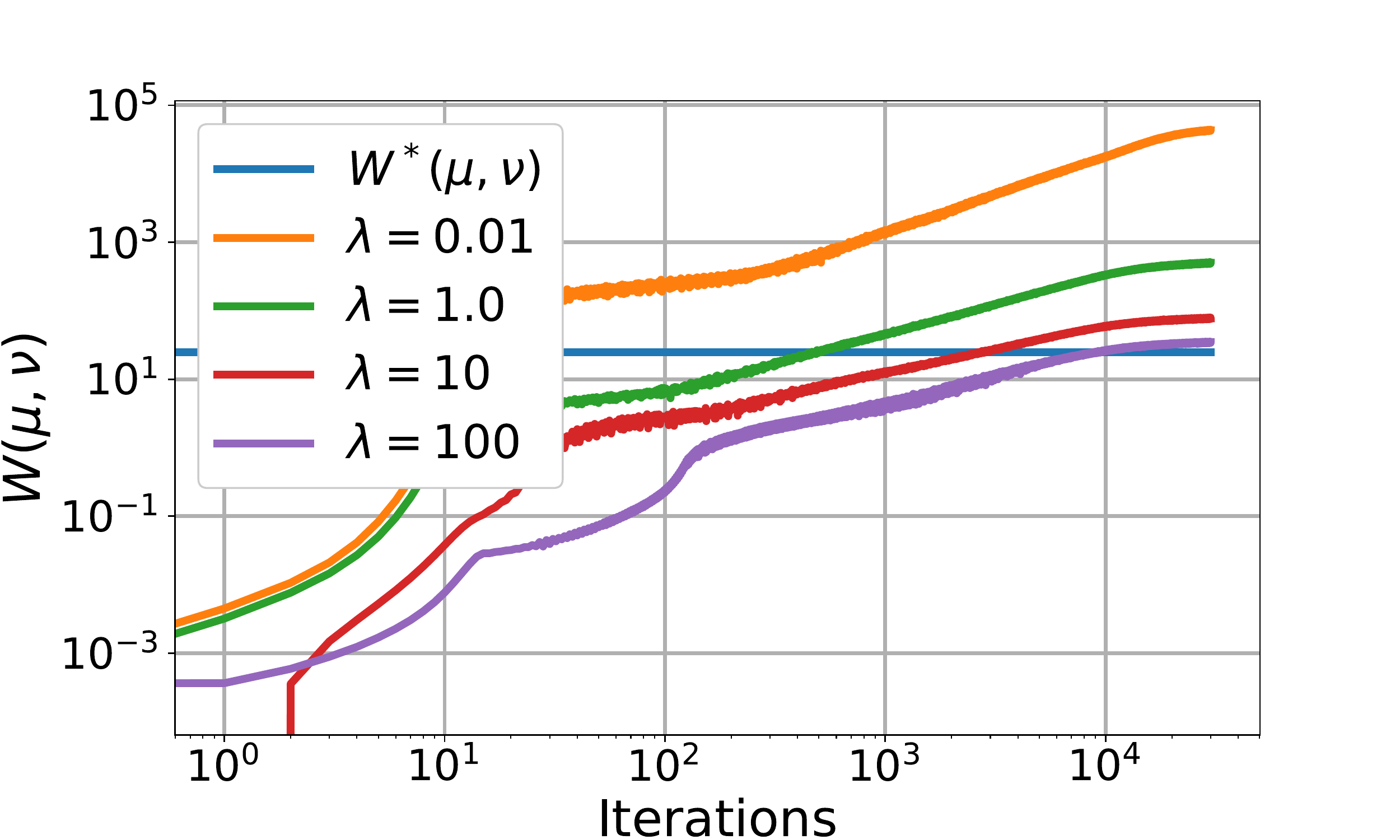}
		\caption{WGAN-GP: Adam $lr=10^{-4}$}
	\end{subfigure}
	\begin{subfigure}[b]{0.48\textwidth}
		\centering
		\includegraphics[width=\textwidth]{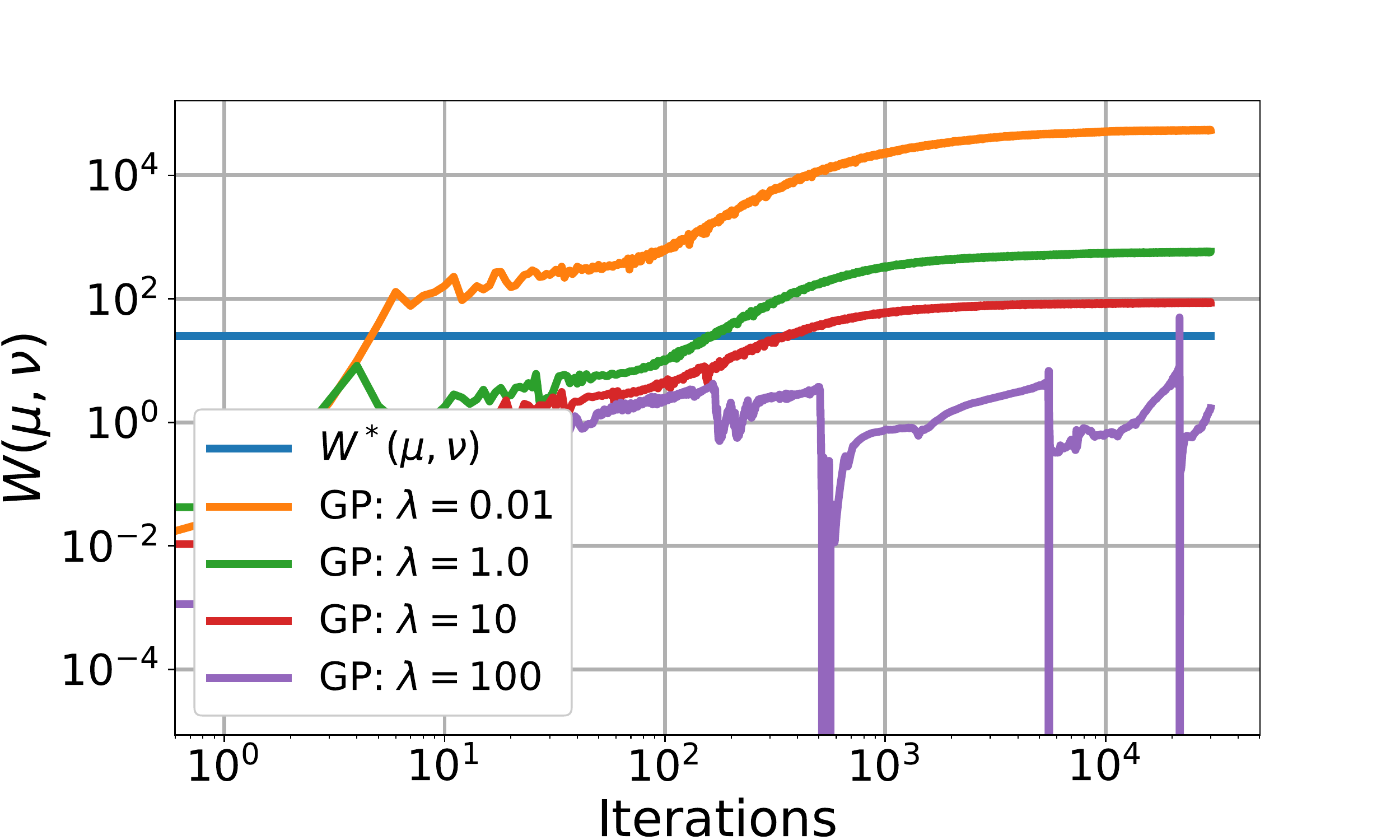}
		\caption{WGAN-GP: Adam $lr=10^{-3}$}
	\end{subfigure}
	 \caption{Optimizer impact on using Gradient Penalty. Notice that the stability changes as the learning rate is changed and the hyperparamter $\lambda$ is changed. Additionally, the slow rate of convergence and the deviation from 1-Lipschitzness are shown here.}
	 \label{fig:gp}
\end{figure}

In this section the hyperparameter dependence of the stochastic algorithms is tested on simple image based examples. For this reason, two batches with size $n=500$ are sampled randomly from the CIFAR dataset and the Wasserstein distance is estimated based on those samples. $n=500$ is used, due to memory restrictions of our GPU and therefore being able to use full batch gradient descent, even for the NN approaches. The results are shown in Fig. \ref{fig:gp}. One can see the stability of the gradient penalty depends on the learning rate of the optimizer and the setting for the lagrangian multiplier $\lambda$. That parameter sensitivity explains the common observation that the Wasserstein estimate heavily oscillates in the initial $10k$ iterations of the generator. Additionally, the estimate has still not converged even after $30k$ full batch iterations. 

\begin{figure}[t]
	\centering
	\begin{subfigure}[b]{0.48\textwidth}
		\centering
		\includegraphics[width=\textwidth]{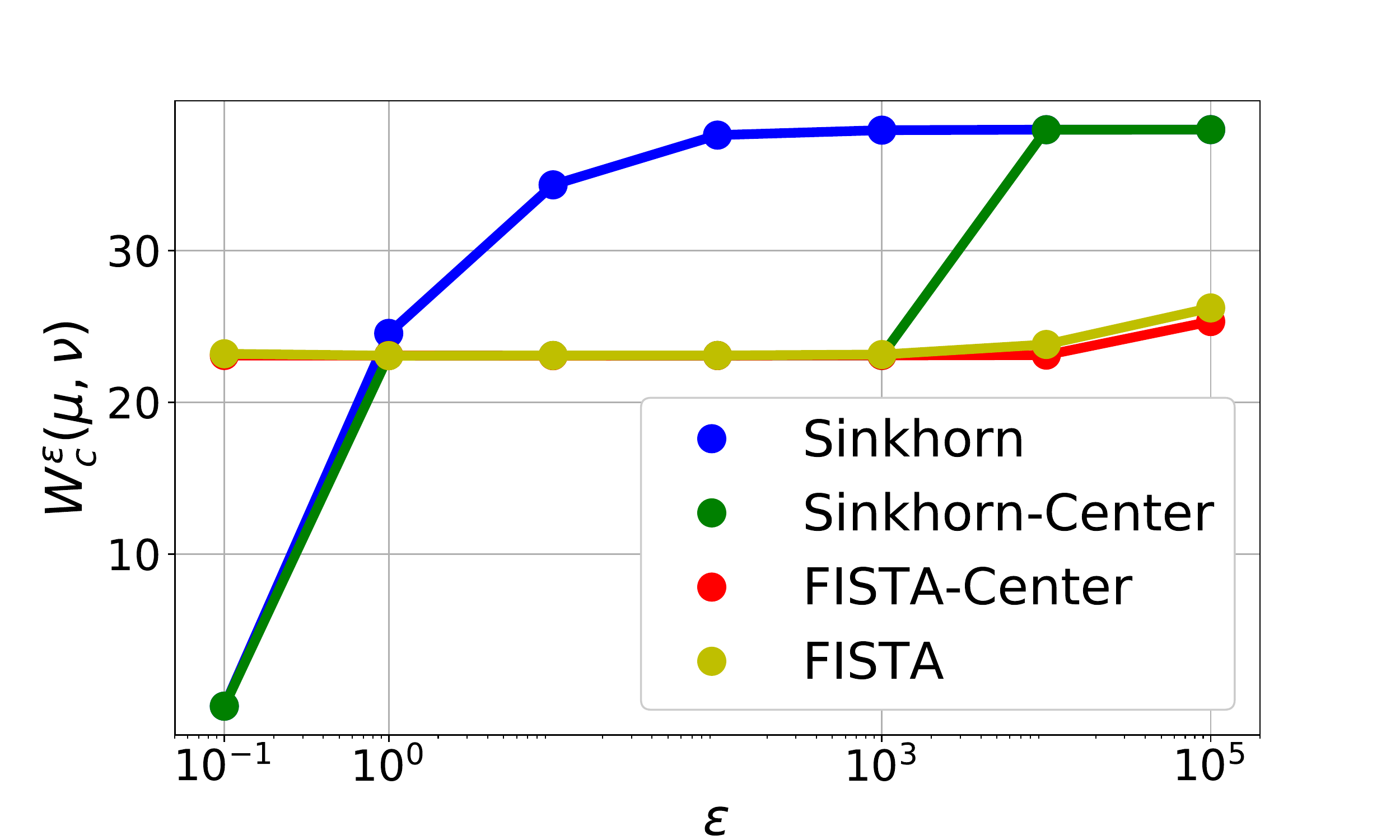}
		\caption{WGAN-GP: Adam $lr=10^{-4}$}
	\end{subfigure}
	\begin{subfigure}[b]{0.48\textwidth}
		\centering
		\includegraphics[width=\textwidth]{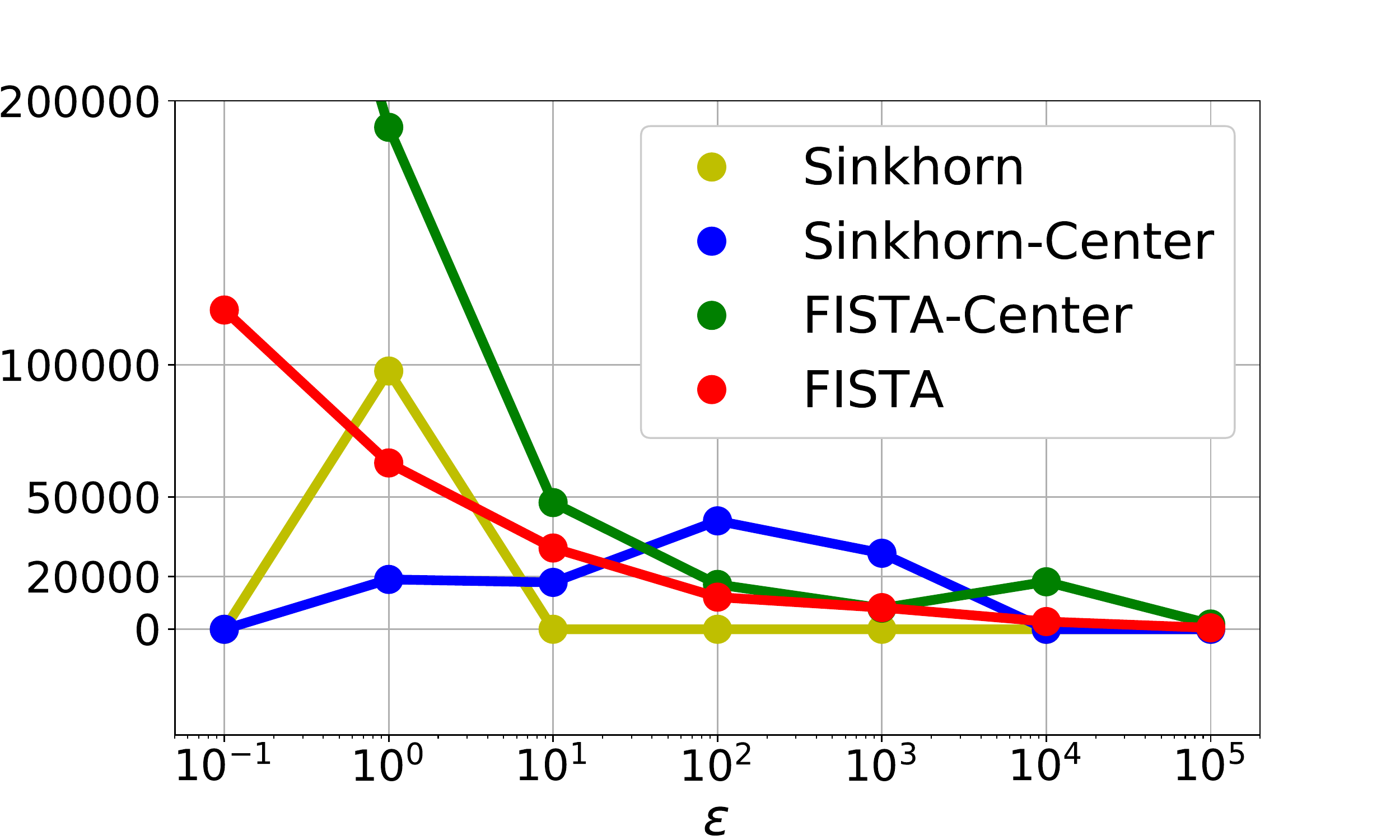}
		\caption{WGAN-GP: Adam $lr=10^{-3}$}
	\end{subfigure}
	\caption{Impact of the regularization term on the estimated Wasserstein distance and the number of iterations until convergence. The stability issues of the Sinkhorn(-Center) algorithms for $0.1 \geq \epsilon$ and $\epsilon \geq 10$ are demonstrated here.}
	\label{fig:reg}
\end{figure}

As a comparison we show the dependence of the fullbatch estimates on their hyperparameter $\epsilon$ in Fig. \ref{fig:reg}. For the entropy regularized versions $\epsilon$ defines a tradeoff between numerical stability and getting good results, where moving the center drastically increases the range of good $\epsilon$ values. On the other hand, for the quadratically regularized versions, $\epsilon$ is a tradeoff between the runtime and the quality of the estimation.

\begin{figure}[t]
	\centering
	\begin{subfigure}[b]{0.48\textwidth}
		\centering
		\includegraphics[width=\textwidth]{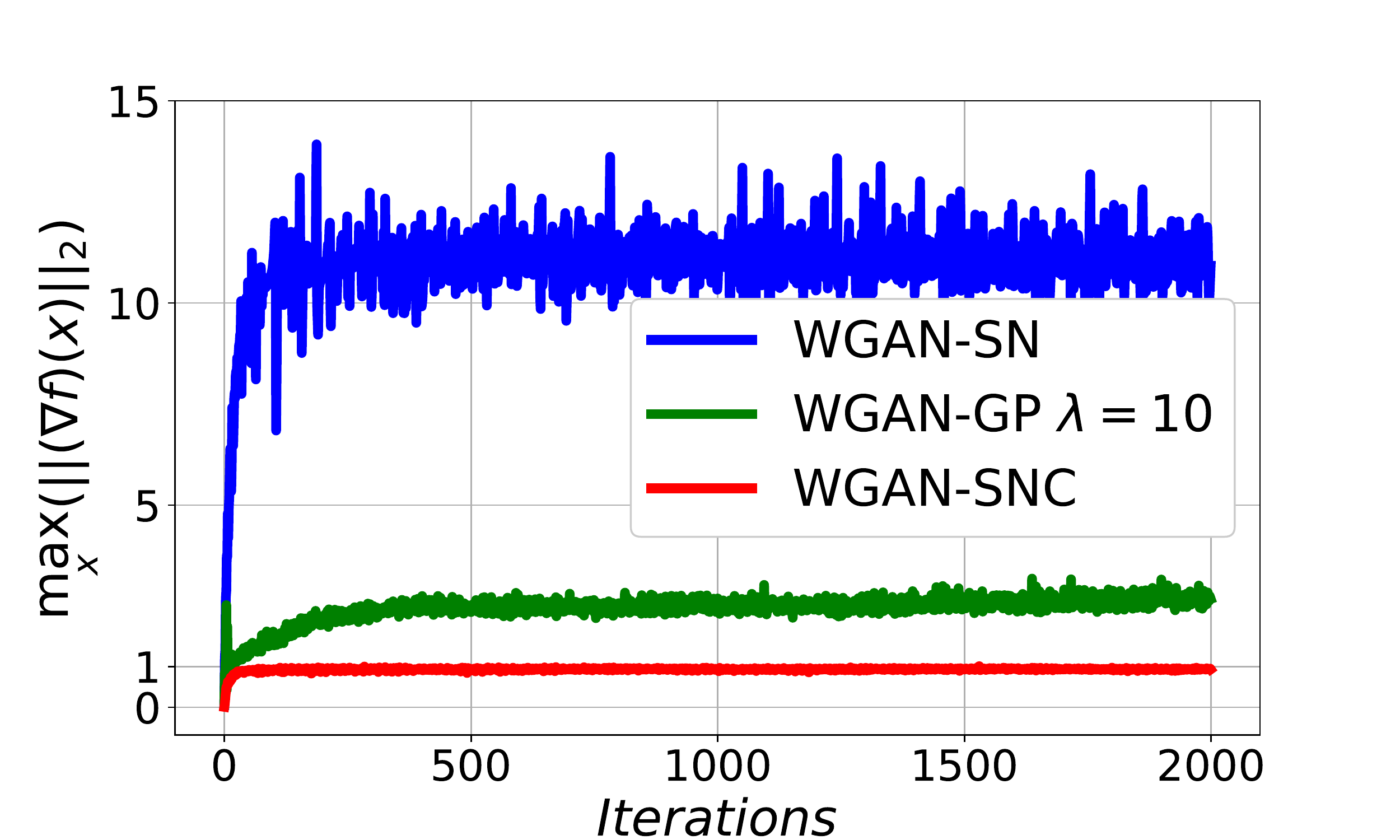}
	\end{subfigure}
	\begin{subfigure}[b]{0.48\textwidth}
		\centering
		\includegraphics[width=\textwidth]{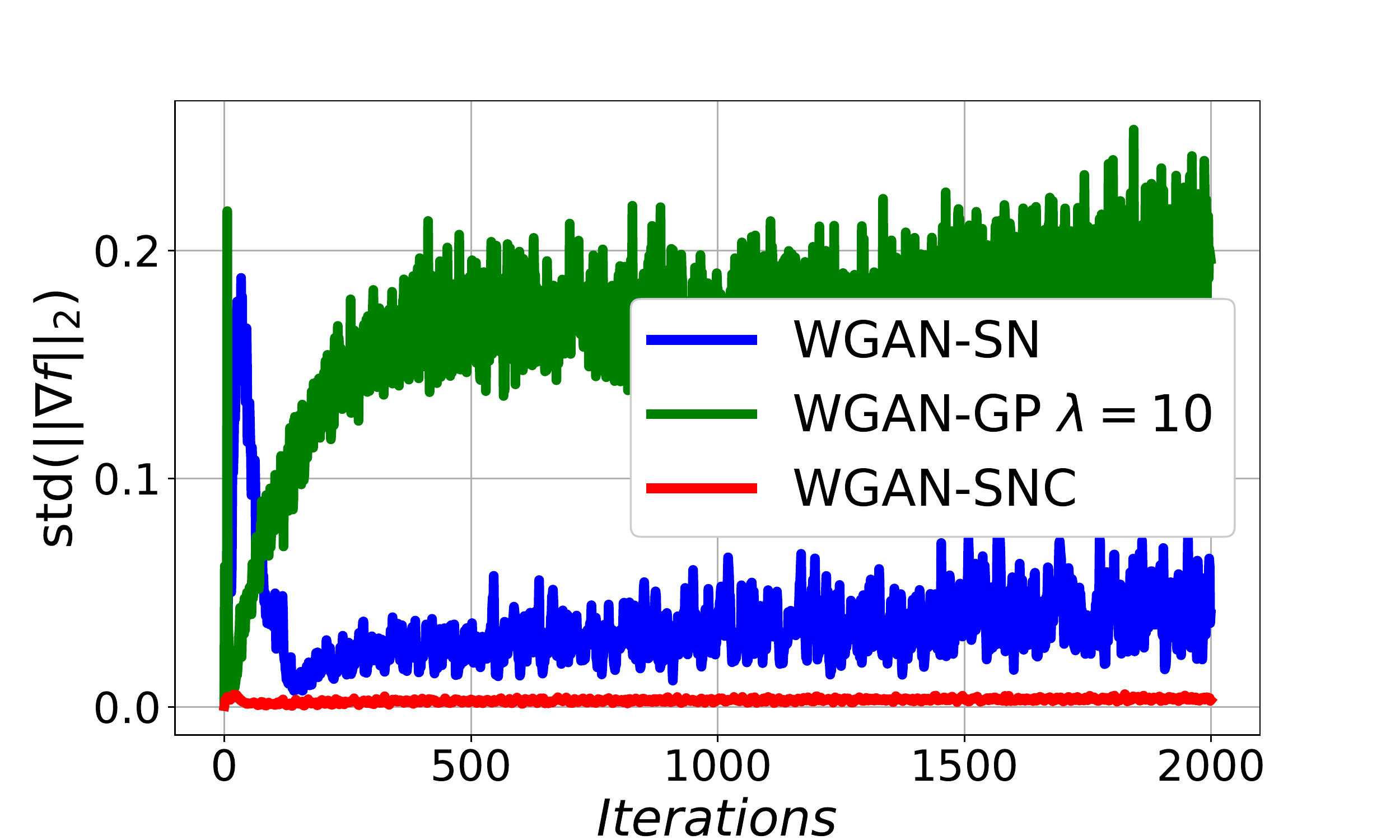}
	\end{subfigure}
	\caption{Empirical Gradnorm of the stochastic algorithms. WGAN-SNC produces stable gradients close to $1$ during training, while the other methods fail to do so.}
	\label{fig:gradnorm}
\end{figure}

\subsection{Limitations of Fullbatch Methods}

\begin{figure}
	\centering
	\begin{subfigure}[b]{0.49\textwidth}
		\centering
		\includegraphics[width=\textwidth]{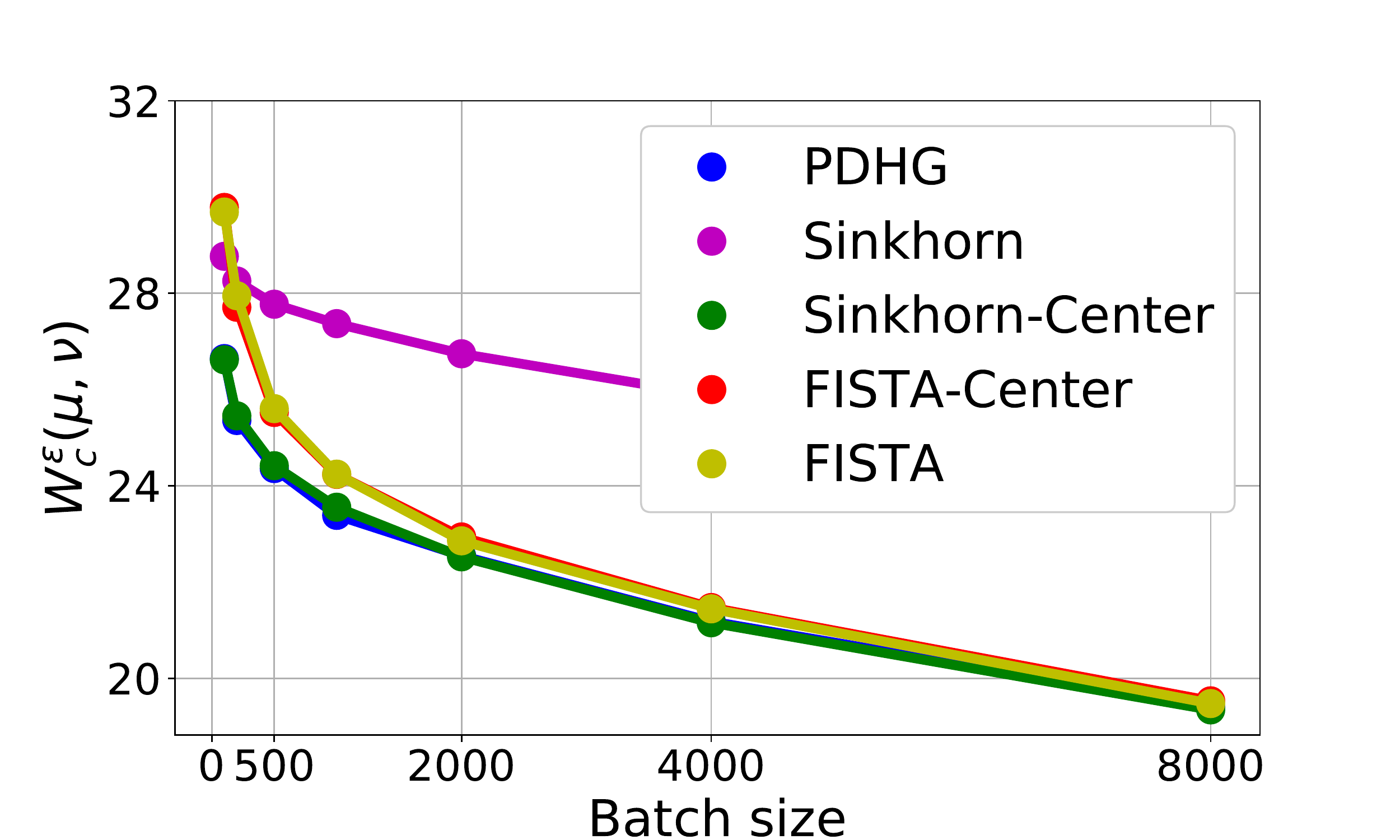}
		\caption{Wasserstein distance}
	\end{subfigure}
	\begin{subfigure}[b]{0.49\textwidth}
		\centering
		\includegraphics[width=\textwidth]{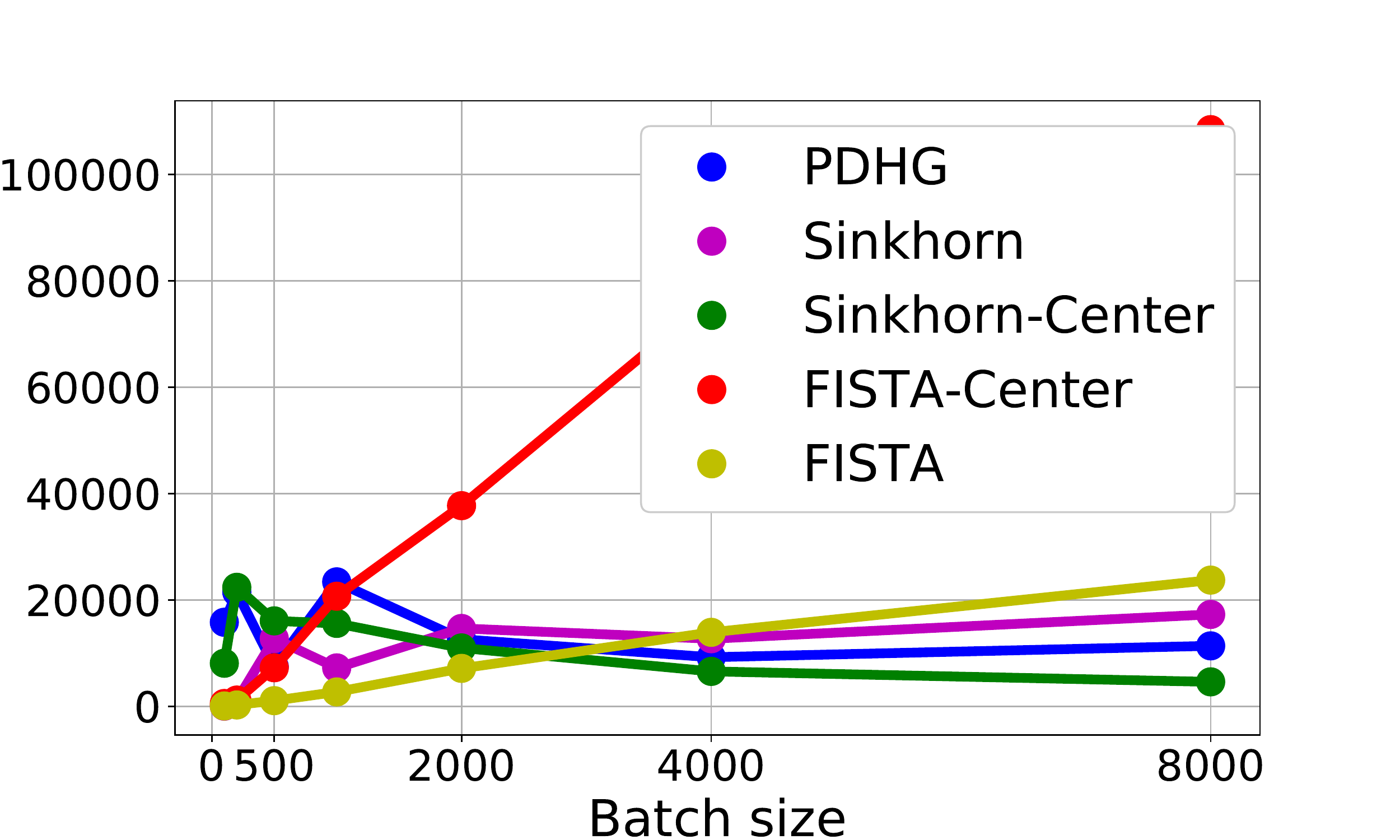}
		\caption{Number of Iterations}
	\end{subfigure}
	\caption{Wasserstein distance and number of iterations until convergence for a specific batch size using L2-norm between random samples on CIFAR.}
	\label{fig:tp}
\end{figure}

In the following two experiments, we show that cost functions without adversarial training do not work well without enormous batch sizes. To showcase, the ability of current generative models, we learn a mapping from a set of noise vectors to a set of images using the Wasserstein distance for a batch of size 4000. The resulting images are shown in the supplementary material for all algorithms. To show that this is not easy to scale to larger datasets, another experiment is designed: the transport map is learned for two different batches taken from CIFAR. The results for an ever increasing number of samples in a batch is shown in Fig. \ref{fig:tp}. This shows an empirical evaluation of the statistical properties of the Wasserstein distance. The estimate decreases with sample size $n$ in the order $O(\sqrt{n})$~\cite{bigot2017central,singh2018minimax}. Even though, those images are sampled from the same distribution, the cost is still higher than using blurred images. Samples produced by the sinkhorn solver in Fig.\ref{fig:cifarsamples} produce an average estimate of $16$ of the Wasserstein distance, which is smaller than samples taken from the dataset itself using the L2-norm. Therefore, it is better for the generator to produce samples like this.

\subsection{Comparison of Algorithms}

Salimans et al.~\cite{salimans2018improving} demonstrated, that with enough computational power it is possible to get state-of-the-art performance using fullbatch methods. However, we do not posses that kind of computational power and therefore the setting proposed by Genevay et al.~\cite{genevay2018learning} was used. We used a standard DCGAN, with a batchsize of $200$, which reproduces their results for the Sinkhorn GAN. For the full batch methods, instead of minimizing the Wasserstein distance, the Sinkhorn divergence\footnote[2]{Sinkhorn divergence $\bar W_{c,\epsilon}(\mu, \nu) = 2W_{c, \epsilon}(\mu, \nu) - W_{c, \epsilon}(\mu, \mu) - W_{c, \epsilon}(\nu, \nu)$} is minimized instead~\cite{genevay2018learning}. The cost function is the mean squared error on an adversarial learned feature space. If the adversarial feature space is only learned on the Wasserstein distance, then features are just pushed away from each other. The Sinkhorn divergence on the other hand also has attractor terms, which forces the network to encode images from the same distribution similarly.

Related GAN algorithms, which also use full batch solvers have been evaluated using the Inception Score~\cite{salimans2016improved} (IS) and a comparison to them is shown in Tab. \ref{tab:algcomp}. For the other methods, we also evaluated using the Fr\'echet Inception Distance (FID)~\cite{heusel2017gans} in Tab. \ref{tab:relwork}. In the constrained setting, the fullbatch WGANs in their current form are competitive or better to similar fullbatch algorithms like the MMD GAN. However, the stochastic methods work better for larger scaled tasks. The WGAN-SN performs better than the one using Gradient Penalty, and performs similarly to the WGAN-SNC. One thing to note is that the WGAN-SN actually works better using a hinge loss, even though no theoretical justification is given for that~\cite{miyato2018spectral}.

\begin{table}[t]
	\begin{adjustwidth}{-1cm}{}
	\centering
	\begin{tabular}{||c | c | c | c | c ||} 
		\hline
		MMD~\cite{genevay2018learning} & Sinkhorn~\cite{genevay2018learning} $\epsilon=10^3$ & Sinkhorn-Center $\epsilon=10^3$ & FISTA $\epsilon=500$ & FISTA-Center $\epsilon=10^4$ \\ [0.5ex] 
		\hline\hline
		$4.04 \pm 0.07$ & $4.10 \pm 0.12$ & $4.69 \pm 0.03$ & $4.25 \pm 0.051$ & $4.61 \pm 0.023$ \\ 
		\hline
	\end{tabular}
	\caption{Inception Score comparison of full batch methods for CIFAR.}
	\label{tab:algcomp}
\end{adjustwidth}
\end{table}

\begin{table}[t]
	\centering
	\begin{tabular}{||c | c | c | c | c | c| c ||} 
		\hline
		- & FISTA-Center & Sinkhorn~\cite{genevay2018learning} & WGAN-GP~\cite{miyato2018spectral} & WGAN SN & WGAN SN conv \\ [0.5ex] 
		\hline\hline
		IS& $4.61 \pm 0.023$ & $4.14 \pm 0.06$ & $6.68 \pm 0.06 $ & $6.80 \pm 0.07$ & $6.93 \pm 0.08$   \\ 
		FID & - & - & $40.1$ & $40.94 \pm 0.55$ & $38.26 \pm 0.68$ \\		
		\hline
	\end{tabular}
	\caption{Visual Quality comparison of Inception Score (IS, higher is better) and FID (lower is better). Note, that the WGAN SN results are our own, as the authors~\cite{miyato2018spectral} did not evaluate the model using the Wasserstein loss.}
	\label{tab:relwork} 
\end{table}

\section{Discussions}

Based on our experimental results, we want to share the following empirical insights to successfully train WGANs for various tasks.

\begin{itemize}
	\item \textbf{Stochastic vs full batch Estimation}: If it is possible to compute the Wasserstein distance for a given problem accurately enough with a full batch approach, then there are a lot of advantages using this approach, like convergence guarantees, better Wasserstein estimates and a clear interpretation of hyperparameters. Depending on the batch size the optimization algorithm might be quite slow though. On the other hand full batch estimation is not possible with simple cost functions for real world image datasets as we show for the CIFAR dataset (see Fig. \ref{fig:cifarsamples}).
	\item \textbf{Cost function:} The cost function controls the geometry of the resulting generated distribution (see Fig. \ref{fig:manifold}). For example the L1-norm results in sharper images and sharper interpolations than the L2-norm~\cite{zhao2017loss} (see Fig. \ref{fig:manifold}).
	\item \textbf{Useful Baseline}: The batch wise estimation gives an indication of the Wasserstein distance given two batches of the dataset. This in turn is used to give an estimate on how well the NN architecture and training algorithm are able to fit the data. (see Fig.\ref{fig:gp})
	\item \textbf{Full batch estimation}
	\begin{itemize}
		\item \textbf{Squared vs Entropy regularization} (see Fig. \ref{fig:tp}): entropy regularization converges extremely fast for large values of $\epsilon$, however, the performance quickly deteriorates even for small changes in $\epsilon$. Quadratic regularization on the other hand works numerically very stable for any value of $\epsilon$ we tested (see Fig. \ref{fig:reg}). Proximal regularization allows for more stability in the algorithm by only minorly changing the algorithm.
		\item \textbf{Batch size:} As a rule of thumb a larger batchsize is better than a smaller one to accurately estimate the Wasserstein distance. Estimating the necessary batchsize is done using indicative batches (e.g. starting batch and batches from the data distribution).
		\item \textbf{Convergence guarantees}: Full batch methods provably converge to the global optimal solution and therefore accurately estimate the Wasserstein distance with a clear meaning of each hyperparameter (see Fig. \ref{fig:reg}). 
	\end{itemize}
	\item \textbf{Stochastic estimation}
	\begin{itemize}
		\item \textbf{Convergence}: In principle, methods based on NNs, take longer to converge, have no convergence guarantees and it is hard to tell, if they really approximate a Wasserstein distance. Additionally, it is unclear how gradients of intermediate approximations relate to a converged approximation, resulting in the mystifying nature of WGAN training.
		\item \textbf{Projection}: Projecting onto the feasible set is too restrictive. Therefore, the projection is done as part of the loss function (see Fig. \ref{fig:toyproj}).
		\item \textbf{Hyperparameter dependence}: Current methods are extremely dependent on hyperparameters (GP on $\lambda$~\cite{lucic2018gans}, and on the optimizer~\cite{miyato2018spectral} and SN on the network architecture) (see Fig. \ref{fig:gp} and Fig. \ref{fig:hyperparameters}).
		\item \textbf{Gradient norm of NNs}: Current methods to ensure Lipschitzness in NNs have in common, that while the actual Lipschitz constant is different from $1$, it is empirically stable. (see Fig. \ref{fig:gradnorm})
	\end{itemize}
\end{itemize}

\section{Conclusions \& Practical Guide}

We have reviewed and extended various algorithms for computing and minimizing the Wasserstein distance between distributions as part of a large generative systems. To make use of those insights in ones problems is to look at the Wasserstein distance between indicative batches, e.g. the initial batches produced by the generator and batches from the data distribution. This also gives a way to gauge how long a NN will take to converge and which hyperparameters have an impact on the estimation. Estimating the Wasserstein distance on indicative batches can safely be done with a regularized solver, due to the small differences in the Wasserstein estimates. For entropy regularization, we encourage to use proximal regularization. If the full batch estimation of the gradient is sufficient, then using a full batch GANs provides reliable results. However, for most GAN benchmarks this is not the case and then Gradient Penalty tends to work well, but is really slow. WGAN-SN is a lot faster, but mathematically incorrect. We propose a theoretically sound version of this, while showing similar performance on CIFAR. The cost function used in the Wasserstein distance controls the geometry of the generated manifold and therefore determines the interpolations between the images. The high cost between different samples taken from the same dataset shows problems with current non-adversarial cost functions on generative tasks and is a first step towards modelling better cost functions. 

	\bibliographystyle{splncs04}
	\bibliography{egbib}

\pagebreak

\section{Supplementary Material}

\subsection{Derivation of the PDHG algorithm}

The initial starting formulation is given as follows:

\begin{equation}
	W_c(\mu, \nu) = \displaystyle \min_{T \in U(\mu, \nu)} \langle T, C \rangle_F,
\end{equation}

By forming the Lagrangian of this formulation and reformulating this yields:
\begin{equation}
\begin{split}
W_c(\mu, \nu) = \min_{T \geq 0}\max_{\lambda_1, \lambda_2} \langle C, T \rangle_F + \lambda_1^T (T\mathbbm{1}_n - \mu) + \lambda_2^T(T^T\mathbbm{1}_m - \nu) \\
W_c(\mu, \nu) = \max_{\lambda_1, \lambda_2} \min_{T \geq 0} \lambda_1^TT\mathbbm{1}_n + \lambda_2^TT^T\mathbbm{1}_m +  \langle C, T \rangle_F - \lambda_1^T\mu - \lambda_2^T\nu \\
\end{split} 
\end{equation}

By reshaping $T, C$ to one dimensional vectors $t, c$ and by combining the constraints $T\mathbbm{1}_n$, $T^T\mathbbm{1}_m$ to $Kt$ the following saddle point formulation is obtained:

\begin{equation}
W_c(\mu, \nu) = \max_{\lambda_1, \lambda_2} \min_{t \geq 0} \langle \begin{bmatrix}\lambda_1 \\ \lambda_2 \end{bmatrix}, Kt \rangle_F + c^Tt - \lambda_1^T\mu - \lambda_2^T\nu 
\end{equation}

 The iterates are given in Alg. \ref{alg:pdhg}.

\subsection{Derivation of quadratic regularization}

The following problem is solved in this section:

\begin{equation}
\min_{T \in U(\mu, \nu)} \langle T, C \rangle_F  + \frac \epsilon 2 ||T||_2^2 
\end{equation}

To solve this we make use of the dual formulation as shown by Blondel et al.~\cite{blondel2017smooth} as follows:

\begin{equation}
W_c^\epsilon = \max_{\alpha, \beta} \alpha^T \mu + \beta^T \nu - \sum_j \sup_{t_j \geq 0} \langle  t_j, \alpha + \beta_j \mathbbm{1}  - C_j \rangle - \epsilon h(t_j)  	
\label{eq:blondelform}
\end{equation}

They also showed that by using the convex conjugate this is reformulated to the following dual problem:

\begin{equation}
W_{c,\epsilon} = \max_{\alpha, \beta} \alpha^T \mu + \beta^T \nu - \frac 1 {2\epsilon} \sum_{ij} [\alpha_i + \beta_j - C_{ij}]_+^2.
\end{equation}

To solve for $\alpha, \beta$, we propose to use Alg. \ref{alg:fista}.

\subsection{Calculation of FISTA-CENTER}

The following problem is solved in this section:

\begin{equation}
T^{k+1} = \arg \min_{T \in U(\mu, \nu)} \langle C, T \rangle_F + D_h(T, T^k), 
\end{equation}

for $h(x) = ||x||^2$. This is simplified by the next sequence of equations:

\begin{equation}
\begin{split}
T^{k+1} = \arg \min_{T \in U(\mu, \nu)} \langle C, T \rangle_F + ||T||^2 - ||T^k||^2 - \langle 2T^k, T - T^k \rangle_F \\
T^{k+1} = \arg \min_{T \in U(\mu, \nu)} \langle C, T \rangle_F + ||T||^2 - ||T^k||^2 - 2\langle T, T^k \rangle_F + 2||T^k||^2 \\
T^{k+1} = \arg \min_{T \in U(\mu, \nu)} \langle C, T \rangle_F + ||T||^2 + ||T^k||^2 - 2\langle T, T^k \rangle_F \\
T^{k+1} = \arg \min_{T \in U(\mu, \nu)} \langle C, T \rangle_F + ||T - T^k||^2
\end{split}
\end{equation}

Plugging this back into the formulation shown in Eq. (\ref{eq:blondelform}) yields:

\begin{equation}
W_c^\epsilon = \max_{\alpha, \beta} \alpha^T \mu + \beta^T \nu - \sum_j \sup_{t_j \geq 0} \langle  t_j, \alpha + \beta_j \mathbbm{1}  - C_j \rangle - \frac \epsilon 2 ||t_j - t^k_j||_2^2  	
\end{equation}

By solving for $T$ analytically, the following result is obtained:

\begin{equation}
t_j = (\frac 1 \epsilon (\alpha + \beta_j \mathbbm{1}  - C_j) +  t^k_j)_+ 
\end{equation}

Plugging this back into the initial equation and setting $x_j = (\alpha + \beta_j \mathbbm{1}  - C_j)$ finally results to:

\begin{equation}
W_c^\epsilon =  \max_{\alpha, \beta} \alpha^T \mu + \beta^T \nu - \sum_j \begin{cases} 
\frac 1 {2\epsilon} ||x_j||^2 +  \langle t^k_j, x_j \rangle \quad \text{if} \ \frac 1 \epsilon x_j +  t^k_j > 0 \\
\frac \epsilon 2 ||t^k_j|| \quad \text{else}
\end{cases}
\end{equation}

Here, the gradient with respect to $\alpha$ is simply:

\begin{equation}
\nabla_\alpha W_c^\epsilon(\mu, \nu) =  \mu - \sum_j \max( \frac 1 \epsilon x_j +  t^k_j, 0)
\end{equation}

The gradient with respect to $\beta$ is given in a similar fashion. Using the gradient, we apply the FISTA algorithm again to solve for $(\alpha, \beta)$. The new transport plan for this algorithm is then given by $(\frac 1 \epsilon (\alpha + \beta_j \mathbbm{1}  - C_j) +  t^k_j)_+$. This way it is possible to improve on the solution of the squared regularization for a given $\epsilon$, by changing the center of the algorithm. Also the algorithm is easily implemented in current deep learning frameworks like tensorflow.

\subsection{Spectral Norm Regularization}

\begin{figure}[t]
	\centering
	\begin{subfigure}[b]{0.48\textwidth}
		\centering
		\includegraphics[width=\textwidth]{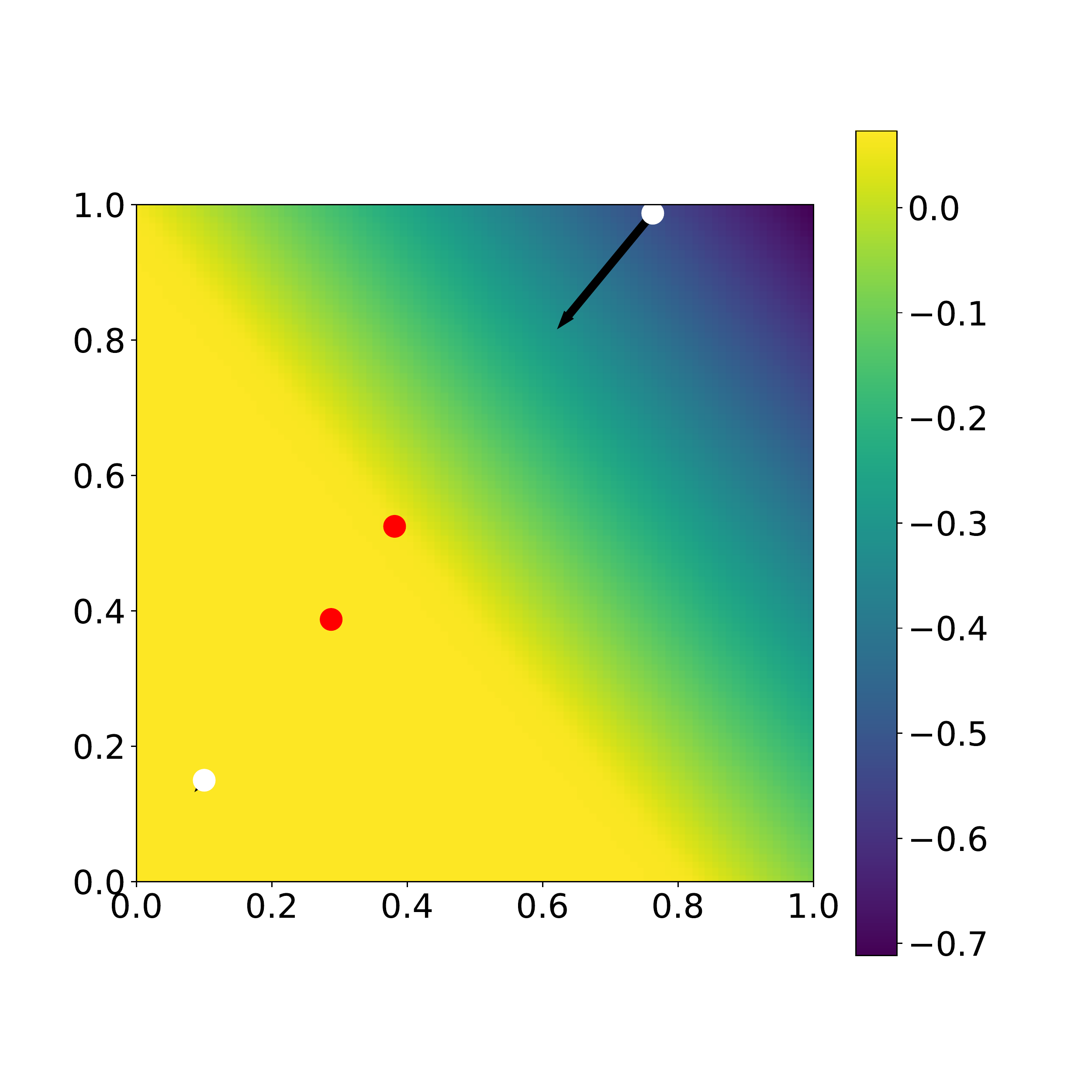}
		\caption{Projection}
	\end{subfigure}
	\begin{subfigure}[b]{0.48\textwidth}
		\centering
		\includegraphics[width=\textwidth]{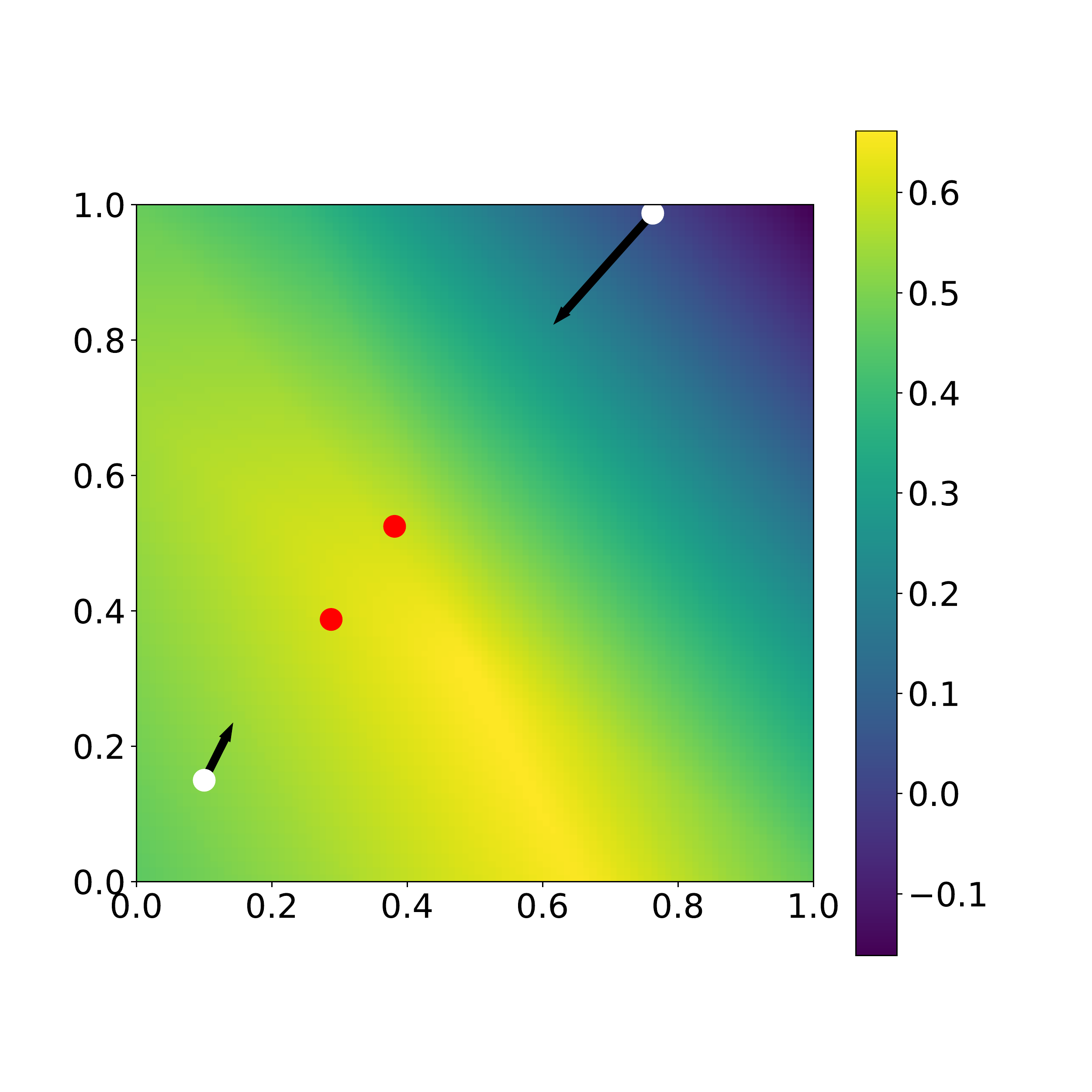}
		\caption{Projection Layer}
	\end{subfigure}
	\caption{Here, the setting encapsulates two data points in red and two generated points in white. A NN with 4 hidden layers each with 10 neurons and ReLU activation was trained using gradient descent to calculate the Wasserstein distance, between those points. The function values of the NNs on this space are visualized and the corresponding gradient used by a generator network to learn where to move those points. As is demonstrated in the figure by using a projected gradient descent the NN gets stuck in a local minima and is unable to learn the non-linear function required.}
	\label{fig:toyproj}
\end{figure}

This is in contrast to the actual SN-GAN~\cite{miyato2018spectral}, where the derivative is calculated through the algorithm. In the context of the Wasserstein GAN, however Fig. \ref{fig:toyproj} demonstrates that the projection approach empirically does not even work for the simplest examples. 

\begin{figure}[t]
	\centering
	\begin{subfigure}[b]{0.48\textwidth}
		\centering
		\includegraphics[width=\textwidth]{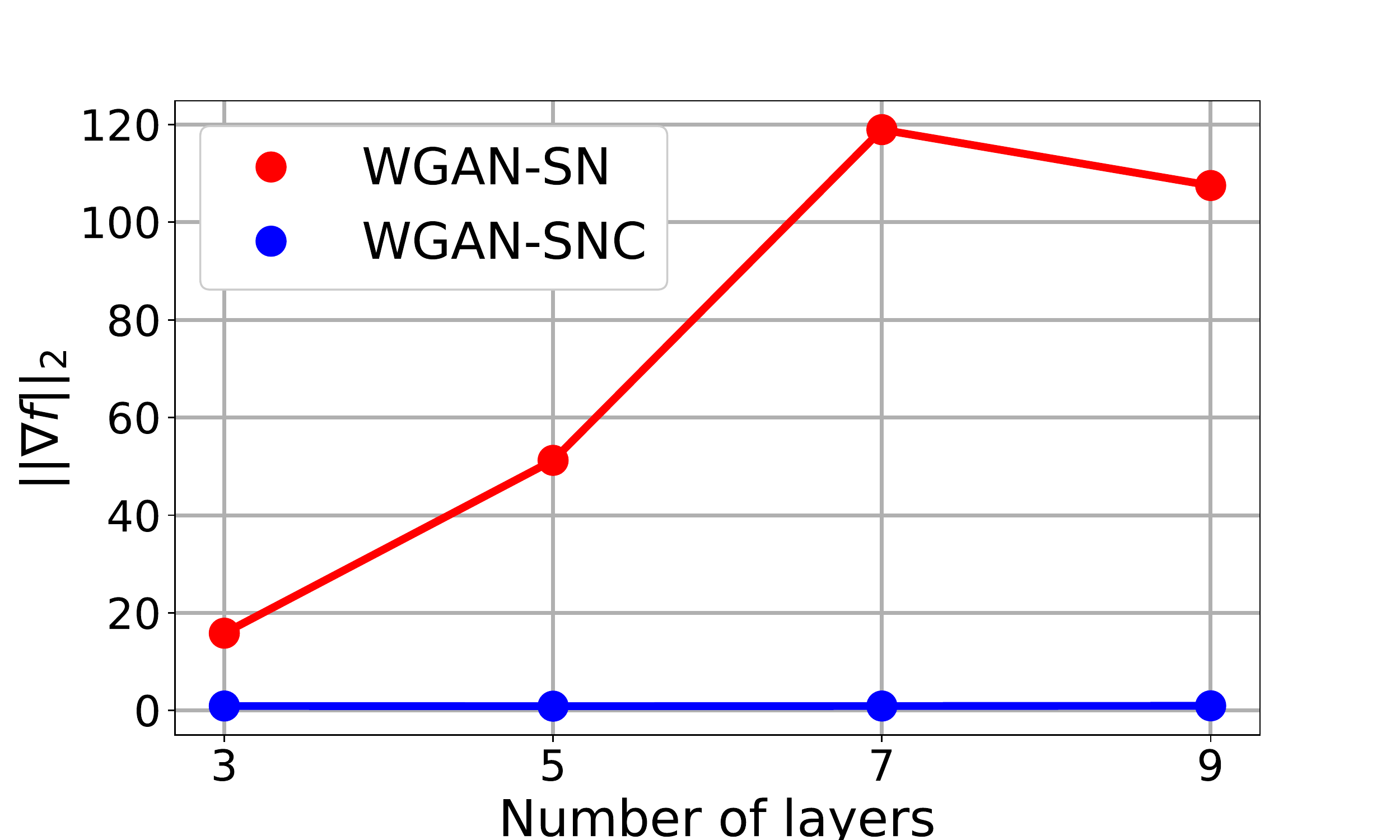}
		\caption{Increasing Layer Numbers}
	\end{subfigure}
	\begin{subfigure}[b]{0.48\textwidth}
		\centering
		\includegraphics[width=\textwidth]{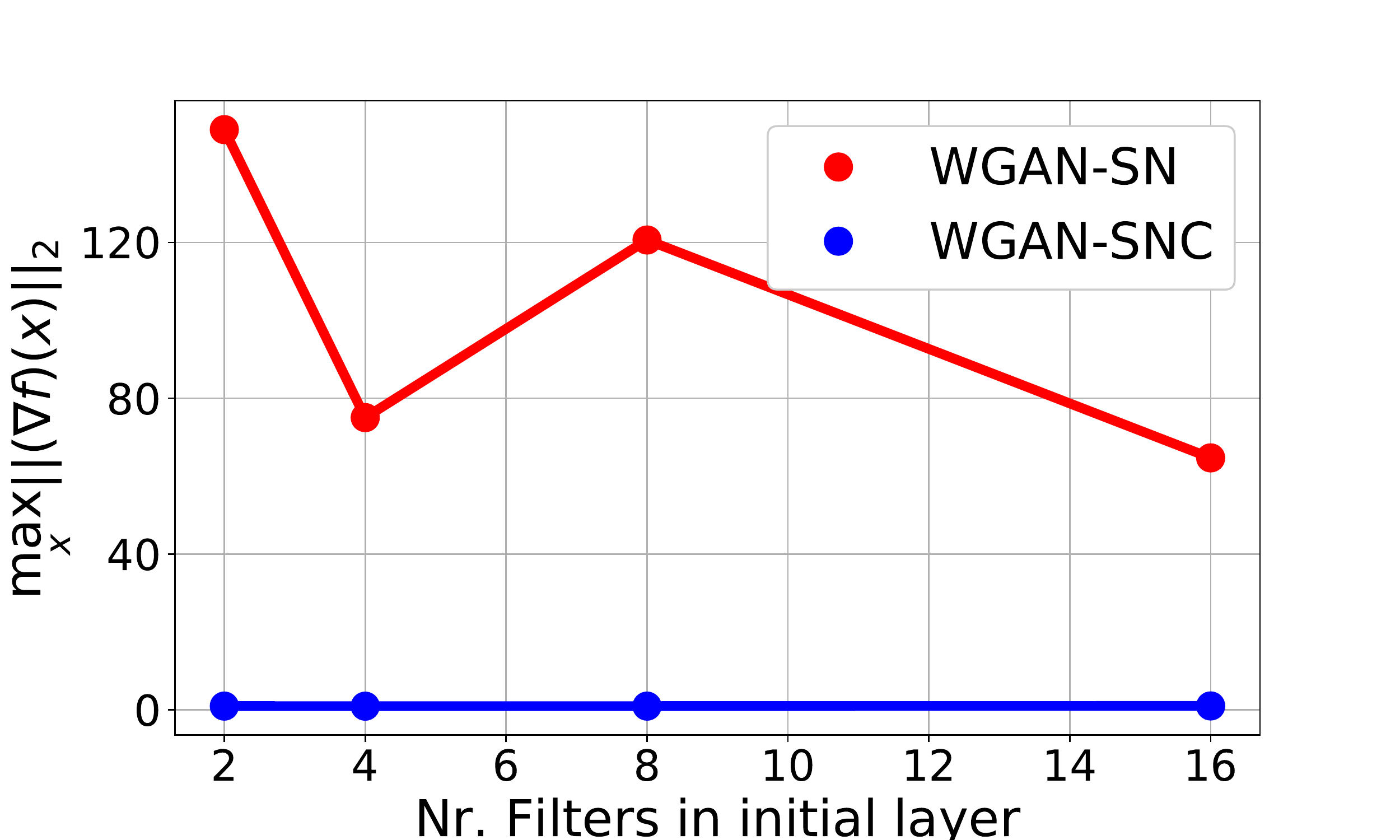}
		\caption{Increasing Feature Maps}
	\end{subfigure}
	\caption{Gradient norm of points in space of a NN regularized by the convolutional spectral norm (WGAN-SNC) and the matrix spectral norm (WGAN-SN~\cite{miyato2018spectral}) calculating the Wasserstein distance between two batches taken from the CIFAR dataset. The desired behaviour would be for those norms to be equal to $1$. Notice that the matrix spectral norm is incorrect and the grad norm is $> 1$.}
	\label{fig:hyperparameters}
\end{figure}

Fig. \ref{fig:hyperparameters} is a demonstration, that the Lipschitz constant of the WGAN-SN increases with the depth of the network, but not with the filter kernels. That the performance of the discriminator increases as you increase the filter weight is shown in the experimental results by Miyato et al.~\cite{miyato2018spectral} to demonstrate the wide applicability of their algorithm.

\subsection{Comparison Image Reconstruction Quality}

\begin{figure}
	\centering
	\begin{subfigure}[b]{0.48\textwidth}
		\centering
		\includegraphics[width=0.95\textwidth]{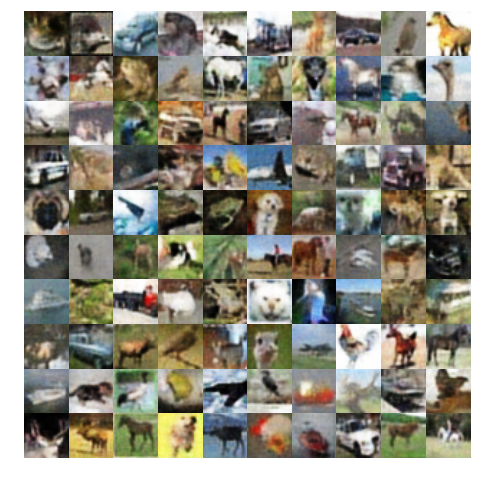}
		\caption{FISTA-Center $d(x,y)=1.57$}
	\end{subfigure}
	\begin{subfigure}[b]{0.48\textwidth}
		\centering
		\includegraphics[width=0.95\textwidth]{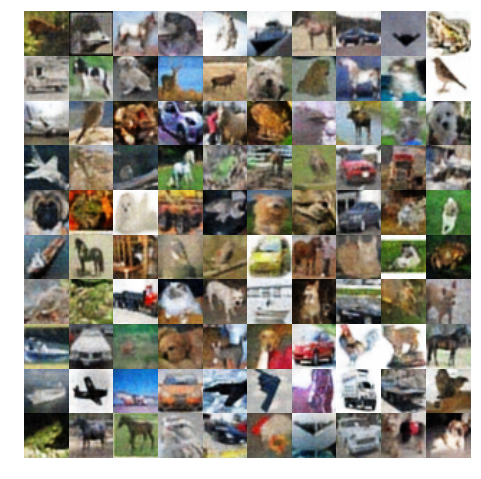}
		\caption{Sinkhorn-Center $d(x,y)=1.98$}
	\end{subfigure}
	
	\begin{subfigure}[b]{0.48\textwidth}
		\centering
		\includegraphics[width=0.9\textwidth]{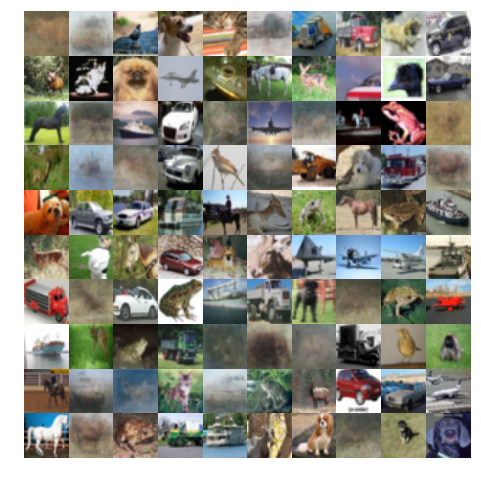}
		\caption{Sinkhorn $d(x,y)=1.42$}
	\end{subfigure}
	\begin{subfigure}[b]{0.48\textwidth}
		\centering
		\includegraphics[width=0.9\textwidth]{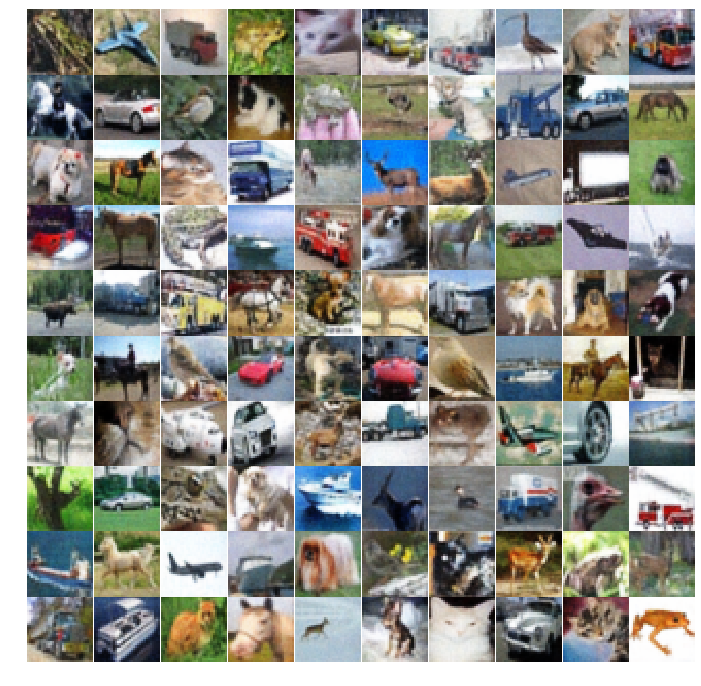}
		\caption{FISTA: $d(x,y)=2.98$}
	\end{subfigure}
	
	\begin{subfigure}[b]{0.48\textwidth}
		\centering
		\includegraphics[width=0.9\textwidth]{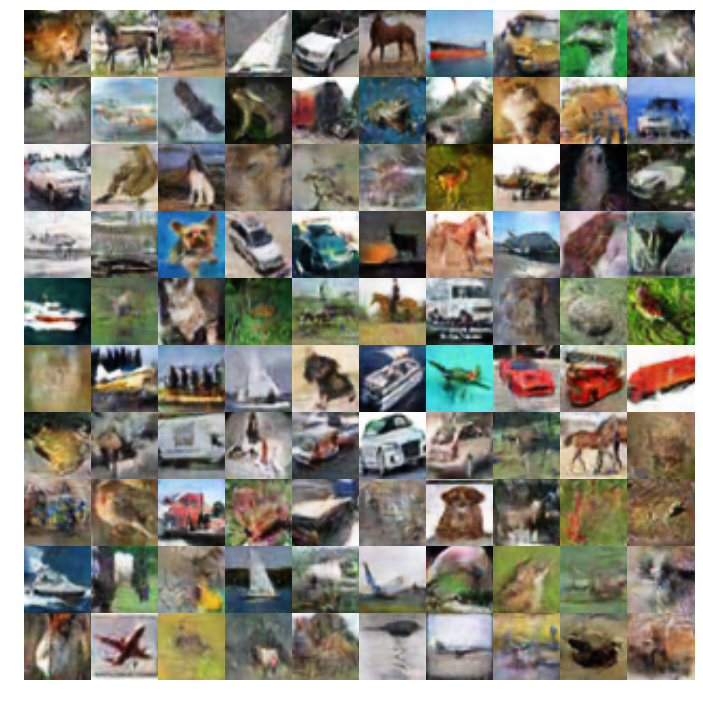}
		\caption{WGAN GP: $d(x,y)=5.46$}
	\end{subfigure}
	\begin{subfigure}[b]{0.48\textwidth}
		\centering
		\includegraphics[width=0.9\textwidth]{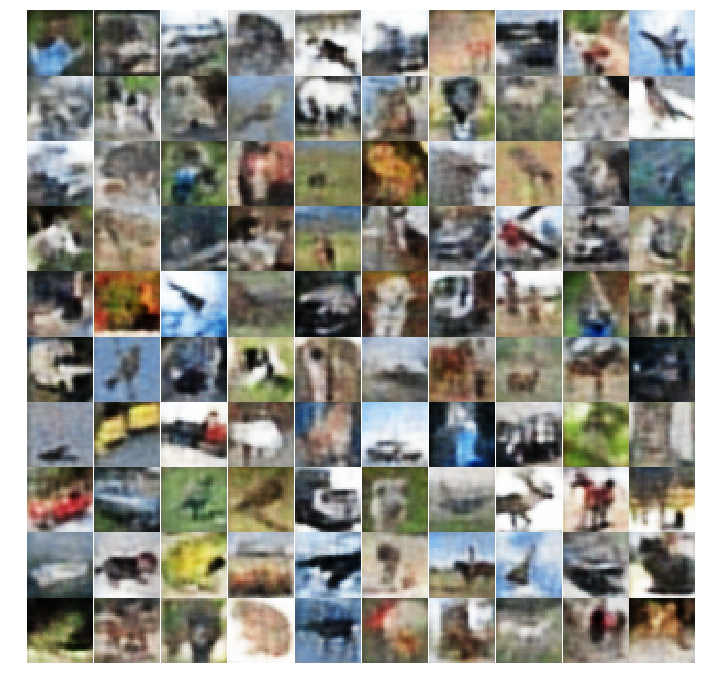}
		\caption{WGAN SN: $d(x,y)=18.22$}
	\end{subfigure}
	\caption{A fixed set of noise samples is used to generate the dataset.}
	\label{fig:cifarsamples}
\end{figure}

In Fig. \ref{fig:cifarsamples} we show the reconstruction capabilities of our full batch algorithms. Therein, $4000$ are reconstructed using the GAN algorithm with a fixed noise set. The estimated Wasserstein distance is then used to learn to reconstruct this dataset using a standard NN. 

\begin{figure}[t]
	\centering
	\includegraphics[width=0.95\textwidth]{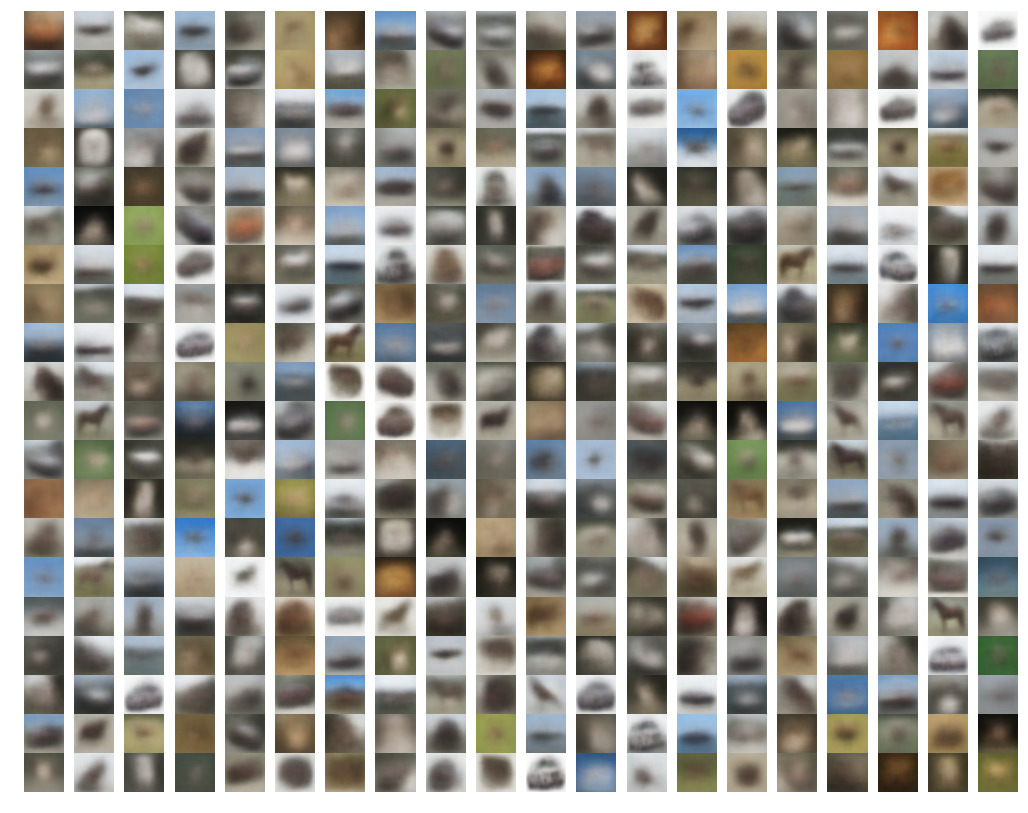}

	\caption{Sample images of the Sinkhorn solver on CIFAR using the L2-norm. The Wasserstein distance here is}
	\label{fig:ipotsamples}
\end{figure}

In Fig. \ref{fig:ipotsamples} we show the failure mode of full batch methods, if the batch size is insufficient for the cost function, which in this case is $1000$ and L2-norm. In Section \ref{sec:exp} in Fig. \ref{fig:tp} it is shown that the Wasserstein distance using the L2-norm is $>20$, while those images produce a Wasserstein distance of $16$.

\subsection{List of Algorithms}

\begin{algorithm}
	\SetAlgoLined
	\KwResult{The final transport map: $t^n$}
	$t^0 = \lambda_1^0 = \lambda_2^0=0$; \\
	$\sigma = \frac 1 {\tau L^2}$ \\
	\While{$k < n$}{
		$t^{k + \tfrac 1 2} = t^k - \tau (K^T \begin{bmatrix} \lambda_1^k \\ \lambda_2^k \end{bmatrix} + c)$ \\
		$t^{k + 1} = \max(t^{k + \tfrac 1 2}, 0)$ \\
		$\hat t^{k+1} = t^{k+1} + (t^{k + 1} - t^k)$ \\
		$\begin{bmatrix} \lambda_1^{k + 1} \\ \lambda_2^{k + 1} \end{bmatrix} = \begin{bmatrix} \lambda_1^k \\ \lambda_2^k \end{bmatrix} + \sigma (K\hat t^{k + 1} - \begin{bmatrix} \mu \\ \nu \end{bmatrix})$ \\
		
	}
	\caption{PDHG: Requires scaling factor $\tau$, with default value $1$ and the cost vector $c$.}
	\label{alg:pdhg}
\end{algorithm}

\begin{algorithm}
	\SetAlgoLined
	\KwResult{The final transport map: $T = \text{diag}(b_n)K\text{diag}(a_n)$}
	$b_0 = 0$; \\
	$K = e^{-\tfrac C \epsilon}$ \\
	\For{k = $0, ...,  n$}{
		$a_{k+1} = \frac \mu {K^Tb_k}$ \\ 
		$b_{k+1} = \frac \nu {Ka_{k+1}}$		
	}
	\caption{Sinkhorn: Requires the cost matrix $C$ and the regularization factor $\epsilon$.}
	\label{alg:sinkhorn}
\end{algorithm}

\begin{algorithm}
	\SetAlgoLined
	\KwResult{The final transport map: $T_n$}
	$b_0 = 0$; \\
	$K = e^{-\tfrac C \epsilon}$ \\
	\For{k = $0, ...,  n$}{
		$Q = T_k \cdot K$ \\
		$a_{k+1} = \frac \mu {Q^Tb_k}$ \\ 
		$b_{k+1} = \frac \nu {Qa_{k+1}}$ \\
		$T_{k+1} = \text{diag}(b_{k+1})Q\text{diag}(a_{k+1})$		
	}
	\caption{Sinkhorn-Center: Requires the cost matrix $C$ and the regularization factor $\epsilon$.}
	\label{alg:sinkhorncenter}
\end{algorithm}

\begin{algorithm}
	\SetAlgoLined
	\KwResult{The final transport map: $T$ with $T_{ij} = \frac 1 \epsilon (\alpha_i + \beta_j - C_{ij})_+$}
	$\alpha^0 = \beta^0 = 0$; \\
	$L = \frac {n + m} \epsilon$ \\
	\For{k = $1, ...,  n$}{
		$\begin{bmatrix}\bar \alpha^k \\ \bar \beta^k \end{bmatrix}  = \begin{bmatrix}  \alpha^k \\  \beta^k \end{bmatrix} + \frac {k - 1} {k + 2} (\begin{bmatrix}  \alpha^k \\  \beta^k \end{bmatrix} - \begin{bmatrix}  \alpha^{k-1} \\  \beta^{k-1} \end{bmatrix})$ \\
		$\alpha^{k + 1}_i = \bar\alpha^k_i + \frac 1 L (\mu_i - \tfrac 1 \epsilon \sum_j(\bar \alpha^k_i + \bar \beta^k_j - C_{ij})_+$ \\
		$\beta^{k + 1}_j = \bar\beta^k_j + \frac 1 L (\nu_j - \tfrac 1 \epsilon \sum_i(\bar \alpha^k_i + \bar \beta^k_j - C_{ij})_+$ \\	
	}
	\caption{FISTA: Requires the cost matrix $C$ and the regularization factor $\epsilon$.}
	\label{alg:fista}
\end{algorithm}

\begin{algorithm}
	\SetAlgoLined
	\KwResult{The final transport map: $T^n$}
	$\alpha^0 = \beta^0 = T^0 = 0$; \\
	$L = \frac {n + m} \epsilon$ \\
	\For{k = $1, ...,  n$}{
		\For{l = $1, ..., n_{\text{inner}}$}{
			$\begin{bmatrix}\bar \alpha^k \\ \bar \beta^k \end{bmatrix}  = \begin{bmatrix}  \alpha^k \\  \beta^k \end{bmatrix} + \frac {k - 1} {k + 2} (\begin{bmatrix}  \alpha^k \\  \beta^k \end{bmatrix} - \begin{bmatrix}  \alpha^{k-1} \\  \beta^{k-1} \end{bmatrix})$ \\
			$\alpha^{k + 1}_i = \bar\alpha^k_i + \frac 1 L (\mu_i -  \sum_j(T^k_{ij} + \tfrac 1 \epsilon (\bar \alpha^k_i + \bar \beta^k_j - C_{ij}))_+$ \\
			$\beta^{k + 1}_j = \bar\beta^k_j + \frac 1 L (\nu_j - \sum_i(T^k_{ij} + \tfrac 1 \epsilon (\bar \alpha^k_i + \bar \beta^k_j - C_{ij}))_+$ \\	
		}
		$T^k_{ij} = \frac 1 \epsilon (\alpha^{n_{\text{inner}}}_i + \beta^{n_{\text{inner}}}_j - C_{ij})_+$
	}
	\caption{FISTA-Center: Requires the cost matrix $C$, number of inner iterations $n_{\text{inner}}$ (default $1000$) and the regularization factor $\epsilon$.}
	\label{alg:fistacenter}
\end{algorithm}

\end{document}